\documentclass[10pt,twocolumn]{article}

\usepackage[a4paper,margin=1.9cm]{geometry}
\usepackage[T1]{fontenc}
\usepackage[utf8]{inputenc}
\usepackage{graphicx}
\usepackage{booktabs}
\usepackage{amsmath,amssymb}
\usepackage{amsfonts}
\usepackage[font=small,labelfont=bf]{caption}
\usepackage{multirow}
\usepackage{array}
\usepackage[table]{xcolor}
\usepackage[numbers,sort&compress]{natbib}
\usepackage[hidelinks]{hyperref}
\usepackage{balance}
\usepackage{dblfloatfix}   

\graphicspath{{figures/}}
\definecolor{artcol}{HTML}{0072B2}
\definecolor{typcol}{HTML}{009E73}
\definecolor{tsacol}{HTML}{D55E00}
\newcommand{\ra}{\ensuremath{\rightarrow}}
\newcommand{\pto}{\textrightarrow}
\newcommand{\art}{\textsc{Art}}
\newcommand{\typ}{\textsc{Typ}}
\newcommand{\asd}{\textsc{ASD}}

\title{\vspace{-1.2em}\bfseries Divergent Gaze Patterns in Artistic Viewing:\\
Spatial \emph{and} Temporal Signatures of Attention Across\\
Autistic Individuals, Artists, and Neurotypical Observers}

\author{
Mohammed Amine Kerkouri$^{1}$ \quad Daphn\'e Senggaran$^{2}$ \quad Renaud Jusiak$^{2}$ \quad Oc\'eane Lehmann$^{2}$\\
Marouane Tliba$^{3}$ \quad Claire Wardak$^{2}$ \quad Emmanuelle Houy-Durand$^{2}$ \quad Shasha Morel-Kohlmeyer$^{2}$\\
Aladine Chetouani$^{4}$ \quad Nadia Aguillon-Hernandez$^{2}$\\[0.4em]
\small $^{1}$F-initiatives, Paris, France \quad $^{2}$Universit\'e de Tours, Tours, France\\
\small $^{3}$Universit\'e d'Orl\'eans, Orl\'eans, France \quad $^{4}$Universit\'e Sorbonne Paris Nord, Villetaneuse, France\\
\small \texttt{m.a.kerkouri@f-initiatives.com}
}
\date{}

\begin{document}
\maketitle

\begin{abstract}
\noindent
How different populations visually explore artworks bears on cognitive
science and on accessibility design, yet most eye-tracking work in autism has
used social scenes rather than art, and has analysed \emph{where} the eyes land
while ignoring \emph{when} and \emph{in what order}. We present a comparative
free-viewing study across three groups, autistic adults (ASD), trained
artists, and neurotypical observers, who each viewed 30 paintings for 15\,s.
We introduce a directed, metric-grounded framework that compares groups along
two complementary axes: a \emph{spatial} axis, in which one group's
fixation-density map predicts another's fixations under six saliency metrics
(AUC-Judd, NSS, CC, SIM, KL, Information Gain); and a \emph{temporal} axis, in
which individual scanpaths are compared with MultiMatch, ScanMatch, a
foveal-disc IoU score (FDISS), and dynamic time warping (DTW). Fixations are
extracted uniformly for all groups with a dispersion-threshold algorithm. Three
results converge. (i)~Artists and neurotypicals are almost indistinguishable in
\emph{both} space (density-map correlation $\text{CC}=0.96$) and time (they form
the most alignable scanpath pair), whereas ASD gaze diverges from both.
(ii)~ASD attention is \emph{dissociated}: it matches artists' wide spatial
exploration (dispersion, explored area) but carries a distinct temporal
signature, shorter fixations, less dwell, and the most idiosyncratic
(least self-consistent) scanpaths of any group. (iii)~ASD gaze is \emph{not}
selectively artist-like on any metric; if anything it is marginally closer to
neurotypical. Together these findings indicate that autistic viewing of art is
a distinct, group-specific attentional profile in both space and time, and they
motivate population-conditioned models of aesthetic attention. We release all
analysis code and per-stimulus results.
\end{abstract}

\section{Introduction}
Paintings are among the richest naturalistic stimuli for studying overt visual
attention. Unlike sparse laboratory displays, an artwork combines low-level
salience, recognisable objects and faces, and deliberate compositional
structure, so that where a viewer chooses to look reflects the interplay of
bottom-up conspicuity, semantic interest, and higher-level viewing strategy
\citep{kerkouri2024gaze,vogt2007expertise}. Free-viewing of artworks therefore
offers a natural setting in which to ask how perceptual expertise, prior
experience, and cognitive style reshape the distribution, and the
dynamics, of gaze.

Two populations are classic ``non-normative'' viewers. \emph{Artists}, trained
to attend to structural and compositional relationships rather than to isolated
objects, distribute gaze more evenly and rely less on the semantically obvious
\citep{vogt2007expertise,francuz2018eye}. \emph{Autistic} individuals (ASD)
show atypical attention to faces and social cues \citep{pelphrey2002visual,
hedley2012using,ricou2025invariant}, heightened engagement with local
low-level features, and reduced influence of global semantic context; these
tendencies are consistent with the weak-central-coherence account
\citep{happe2006weak} and the enhanced-perceptual-functioning model
\citep{mottron2006enhanced}, and free-viewing gaze in ASD is more strongly tied
to pixel-level salience than to object- and semantic-level content
\citep{wang2015atypical}. Yet almost all of this evidence comes from social
scenes and faces; how autistic observers deploy attention during \emph{aesthetic}
viewing of artworks is comparatively unexplored.

This leaves a deceptively simple question. Both artists and autistic observers
attend to detail rather than to the single obvious focal point. Does ASD gaze
during artwork viewing therefore resemble the detail-oriented gaze of trained
artists, the gaze of neurotypical observers, or neither? A preliminary
late-breaking study \citep{kerkouri2024gaze} suggested a dissociation at the
level of spatial density maps, but it was limited in three ways that this paper
addresses: it analysed only spatial density (discarding the temporal order of
fixations), it relied on a single map per group per stimulus, and it was
statistically underpowered.

We make four contributions.
\textbf{(1)~A two-axis, directed, metric-grounded framework} for cross-group
gaze comparison that treats group-to-group similarity both as a
saliency-prediction problem (spatial axis) and as a scanpath-comparison problem
(temporal axis), using a uniform fixation-extraction pipeline applied
identically to every group.
\textbf{(2)~A three-group free-viewing comparison} (artists, neurotypical
observers, autistic adults) over 30 paintings, with fixations re-extracted from
the raw signal, yielding $2{,}091$ scanpaths and $95{,}282$ fixations.
\textbf{(3)~Evidence of a spatial--temporal dissociation in ASD}: autistic gaze
matches artists' wide spatial exploration but has a distinct temporal
signature, and is the most idiosyncratic (least self-consistent) of the three
groups.
\textbf{(4)~A clear ordering}: artists and neurotypicals converge in both space
and time, while ASD is not selectively artist-like on any of ten metrics.

\section{Related Work}
\paragraph{Expertise and gaze in art perception.}
A substantial literature establishes that formal training reshapes how art is
viewed. \citet{vogt2007expertise} found that artists distribute gaze more evenly
across a composition and rely less on the salient objects that dominate novice
viewing, and \citet{francuz2018eye} report converging eye-movement correlates of
visual-arts expertise. Because gaze on artworks is driven by compositional and
stylistic structure that differs from everyday scenes, predicting it
computationally is itself a domain-specific problem, addressed through domain
adaptation \citep{kerkouri2022domain}, self-supervised learning
\citep{tliba2022selfsup}, and dedicated art-viewing datasets
\citep{kerkouri2024avatt}. We build on this line but ask a comparative,
population-level question rather than modelling a single population.

\paragraph{Visual attention in autism.}
Atypical gaze is among the most robust behavioural markers associated with ASD.
Early scanning studies showed autistic individuals fixate faces and social
scenes differently from neurotypical viewers \citep{pelphrey2002visual}, and
eye-movement measures have been used as implicit indices of face processing
\citep{hedley2012using,ricou2025invariant}. Two influential accounts predict a
bias away from global, semantically driven looking: weak central coherence
\citep{happe2006weak} and enhanced perceptual functioning
\citep{mottron2006enhanced}. Directly relevant, \citet{wang2015atypical} used
model-based eye tracking to show that free-viewing gaze in ASD is atypically
driven by low-level pixel salience. Prior work concentrated on social and
naturalistic scenes; free-viewing of paintings, where composition and aesthetics
matter, remains largely unaddressed.

\paragraph{Comparing gaze: distributions and scanpaths.}
For distribution-level comparison we build on the saliency-evaluation
literature: \citet{bylinskii2018metrics} provide a systematic account of what
location-based (AUC, NSS) and distribution-based (CC, SIM, KL) metrics measure,
and \citet{kummerer2015information} introduce an information-theoretic
formulation (Information Gain). Although designed to score models, these metrics
apply to any pair of fixation distributions: treating one group's density map as
the ``model'' and another group's fixations as the ``ground truth'' yields a
principled directional measure of cross-group similarity. The order and timing
of fixations, however, are discarded by density maps. Scanpath-comparison
measures recover them: MultiMatch decomposes similarity into shape, direction,
length, position and duration \citep{dewhurst2012multimatch}; ScanMatch adapts
sequence alignment to binned fixation strings \citep{cristino2010scanmatch};
perceptually grounded overlap can be quantified with foveal discs
\citep{kerkouri2026fdiss}; and semantic similarity can be computed with
vision--language models \citep{kerkouri2026semantic}. We combine both families
into a single directed framework and, crucially, apply it to human populations
rather than algorithms.

\section{Data \& Participants}
Eye-tracking data were recorded with an SMI RED500 remote eye tracker at
500\,Hz while participants freely viewed 30 colour paintings. Each stimulus was
presented at $1650\times1050$\,px for $\approx$15\,s (verified from the
per-trial export timings), with no explicit task, so that recorded gaze reflects
spontaneous exploration. The 30 paintings span landscapes, still lifes,
portraits and abstract compositions. Each participant saw all 30 paintings,
split across two blocks with no repetition, giving one scanpath per participant
per painting. Three groups were recruited:
\begin{itemize}\itemsep2pt
  \item \textbf{Artists} ($n=24$): trained visual artists with $\geq$10 years of
        professional or semi-professional experience.
  \item \textbf{Neurotypical} ($n=32$): no formal art training and no
        neurological diagnosis.
  \item \textbf{ASD} ($n=15$): autistic adults diagnosed per ICD-11
        \citep{harrison2021icd} without co-occurring intellectual disability.
\end{itemize}
After quality control (Sec.~\ref{sec:extract}), the analysis set comprises
$2{,}091$ scanpaths and $95{,}282$ fixations
(\art~712, \typ~940, \asd~439 scanpaths).

\section{Methodology}
We compare groups along two axes. The \emph{spatial} axis
(Sec.~\ref{sec:spatial}) asks how well one group's fixation \emph{distribution}
predicts another's. The \emph{temporal} axis (Sec.~\ref{sec:scanpath}) asks how
similar the ordered fixation \emph{sequences} are. Both consume a single set of
fixations extracted uniformly for every group.

\subsection{Fixation extraction}\label{sec:extract}
The averaged binocular point of regard was median-smoothed (15-sample window,
$\approx$30\,ms) to suppress remote-tracker jitter, then classified into
fixations with the dispersion-threshold identification algorithm (IDT)
\citep{salvucci2000identifying} as implemented in the \texttt{eyefeatures}
library \citep{eyefeatures2024} (maximum dispersion $\approx$90\,px $\approx2^\circ$,
minimum duration 100\,ms). We deliberately did \emph{not} use the tracker's own
per-sample event labels, which on this recording are implausible ($\sim$47\% of
samples labelled saccade); a single dispersion rule applied to all groups avoids
confounding group differences with per-participant classifier behaviour.
Scanpaths with fewer than four fixations were discarded. The procedure recovers
physiological fixations (mean duration $\approx$200\,ms, $\approx$46 fixations
per 15\,s viewing).

\subsection{Spatial axis: density-map prediction}\label{sec:spatial}
For each group $g$ and stimulus $s$ we pool the fixations of all participants in
$g$ and rasterise them, on a down-scaled $210\times131$ grid, into a binary map
$Q^s_g$; smoothing with a 2-D Gaussian ($\sigma\!\approx\!1^\circ$) and
normalising yields a continuous density map $P^s_g$. A directed comparison
$g\!\ra\!h$ uses $P^s_g$ as a pseudo-predictor of group $h$'s fixations, scored
against $h$'s continuous map (distribution metrics CC, SIM, KL) or against $h$'s
fixation locations $\mathcal{F}^s_h$ (location metrics AUC-Judd, NSS,
Information Gain over a centre-prior baseline). We report the stimulus-averaged
score for every ordered pair of distinct groups; CC and SIM are symmetric, the
other four are directional.

Writing $P$ for the (probability-normalised) predictor map, $Q$ for the target's
map, $\mathcal{F}$ for the target's fixation locations, and $B$ for a centred
Gaussian baseline, the six metrics are
\begin{align}
\text{CC}  &= \operatorname{cov}(P,Q)/(\sigma_P\sigma_Q), \\
\text{SIM} &= \textstyle\sum_i \min(P_i,Q_i), \\
\text{KL}  &= \textstyle\sum_i Q_i \log(Q_i/P_i), \\
\text{NSS} &= \tfrac{1}{|\mathcal{F}|}\textstyle\sum_{x\in\mathcal{F}} (P(x)-\mu_P)/\sigma_P, \\
\text{IG}  &= \tfrac{1}{|\mathcal{F}|}\textstyle\sum_{x\in\mathcal{F}} \big[\log_2 P(x)-\log_2 B(x)\big], \\
\text{AUC} &= \text{ROC-area}(P;\mathcal{F}),
\end{align}
where AUC-Judd sweeps a threshold over $P$, treating target fixations as
positives \citep{bylinskii2018metrics,kummerer2015information}. Higher is better
except KL (lower = closer).

\subsection{Temporal axis: scanpath comparison}\label{sec:scanpath}
A scanpath is the ordered sequence of fixations $(x_i,y_i,d_i)$ for one
participant on one painting. For every pair of participants viewing the same
painting we compute four complementary measures:
\textbf{MultiMatch} \citep{dewhurst2012multimatch}, giving five similarity
sub-scores (shape, direction, length, position, duration);
\textbf{ScanMatch} \citep{cristino2010scanmatch}, a Needleman--Wunsch alignment
score over a $12\times8$ spatial grid;
\textbf{FDISS} \citep{kerkouri2026fdiss}, a symmetric foveal-disc IoU score in
$[0,1]$ (foveal radius $=1^\circ$); and
\textbf{DTW}, the dynamic-time-warping distance between the two
$(x,y)$ sequences. Each unordered participant pair is labelled by the sorted
pair of group codes, yielding three within-group cells (\texttt{art-art},
\texttt{typ-typ}, \texttt{tsa-tsa}) and three between-group cells
(\texttt{art-typ}, \texttt{art-tsa}, \texttt{tsa-typ}); this gives $71{,}840$
pairwise comparisons.

\subsection{Per-scanpath features}
To characterise \emph{what} differs, we compute nine features per scanpath:
fixation count, mean fixation duration, total dwell time, mean saccade
amplitude, total scanpath length, spatial dispersion (mean distance of
fixations from their centroid), explored area (convex-hull fraction of the
screen), and the horizontal/vertical centroid of fixations.

\subsection{Statistical analysis}
The primary question, whether ASD gaze is selectively closer to artists or to
neurotypicals, is tested by contrasting the two ASD-source comparisons
(\asd\ra\art\ vs.\ \asd\ra\typ\ spatially; \texttt{art-tsa} vs.\ \texttt{tsa-typ}
temporally), treating the 30 paintings as paired observations with paired
$t$-tests, Cohen's $d$, and post-hoc power. Per-scanpath features are compared
across groups with Kruskal--Wallis omnibus tests and Holm-corrected pairwise
Mann--Whitney tests (rank-biserial effect sizes). All code and per-stimulus
outputs are released with the paper.

\section{Results}
\begin{figure*}[tbp]
  \centering
  \includegraphics[width=0.80\textwidth]{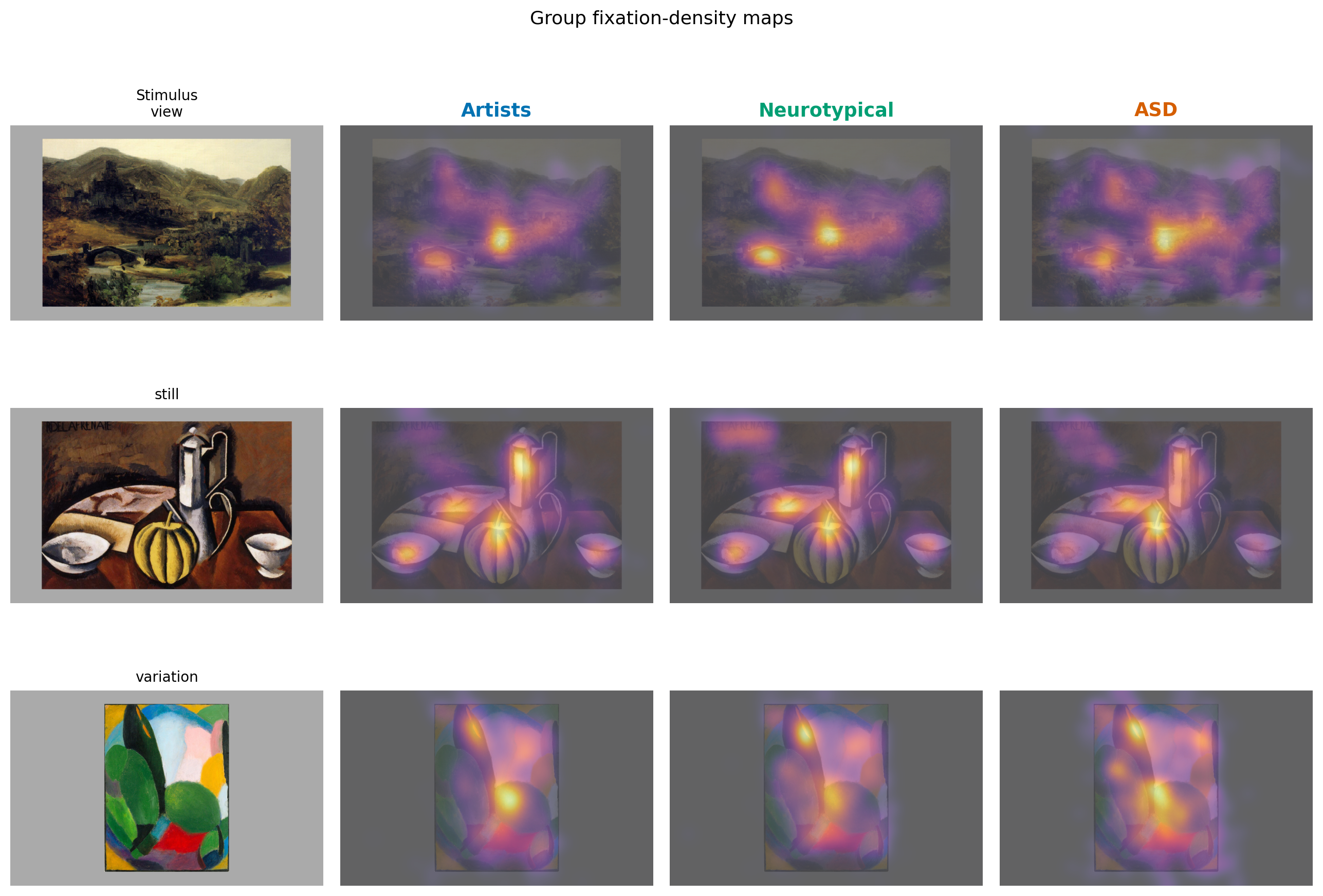}
  \caption{Group fixation-density maps for three example paintings (landscape,
  still life, abstract). Artist and neurotypical maps concentrate on the same
  regions; ASD maps are more diffuse and spatially shifted.}
  \label{fig:density}
\end{figure*}

Figure~\ref{fig:density} shows example group density maps; artist and
neurotypical maps are visibly concentrated on shared regions, whereas ASD maps
are more diffuse and shifted.

\subsection{Spatial axis: artists $\approx$ neurotypicals, ASD distinct}
\begin{figure*}[tbp]
  \centering
  \includegraphics[width=0.96\textwidth]{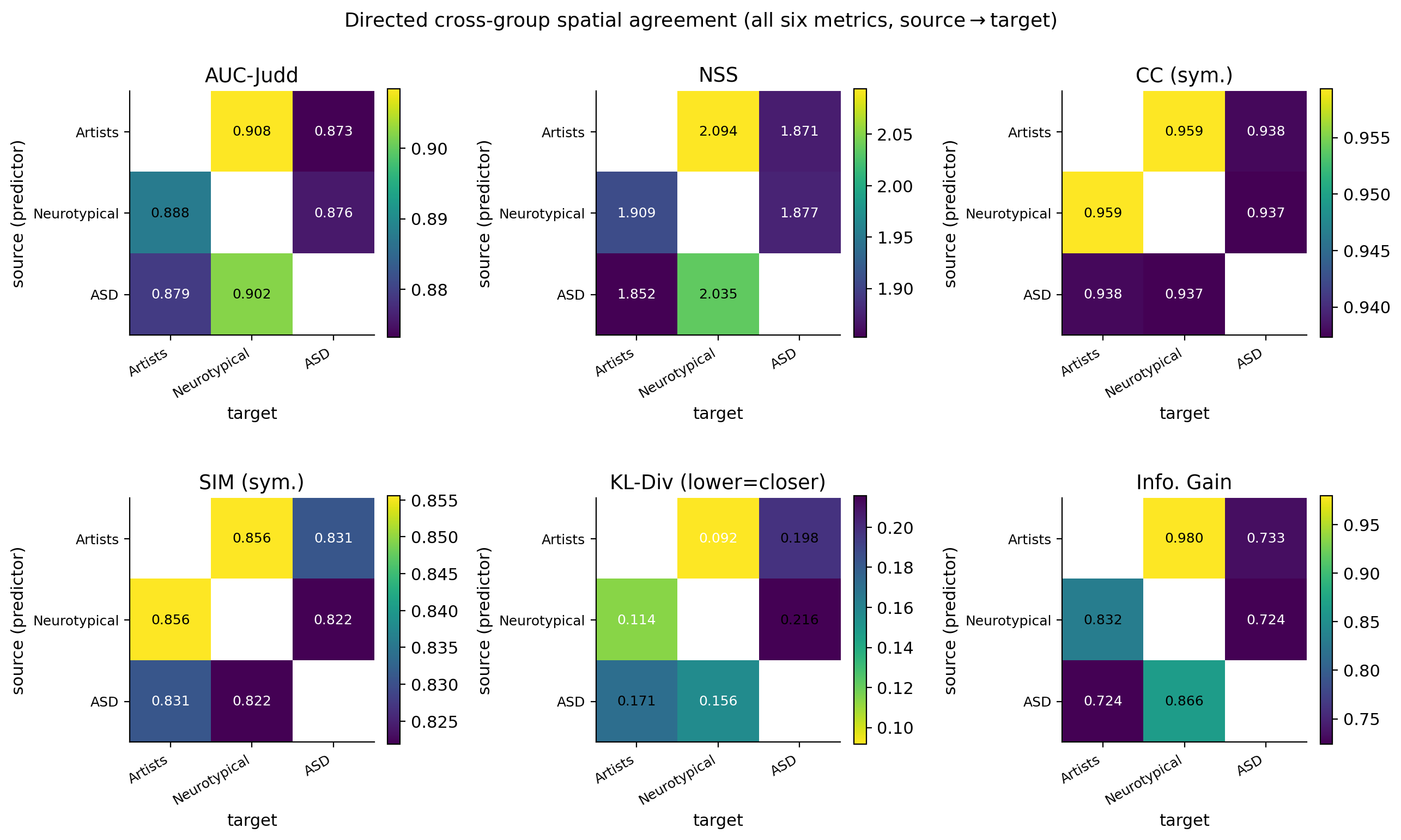}
  \caption{Directed cross-group spatial agreement for all six metrics
  (source\pto target; each cell is a group pair). The symmetric metrics (CC, SIM)
  show the \art--\typ\ cell brightest; the directional metrics (AUC, NSS, IG, and
  KL, plotted with a reversed scale so brighter = closer) show that predicting
  ASD (right column) is hardest.}
  \label{fig:spatialmat}
\end{figure*}

Table~\ref{tab:spatial} and Fig.~\ref{fig:spatialmat} summarise the directed
spatial comparison. Two patterns stand out. First, the symmetric
distribution metrics place artists and neurotypicals almost on top of each other
(\texttt{art--typ} $\text{CC}=0.959$, $\text{SIM}=0.856$), while every pair
involving ASD is lower and nearly identical to one another
(\asd--\art\ $\text{CC}=0.938$; \asd--\typ\ $\text{CC}=0.937$): the ASD density
map is the odd one out, and about equally distant from both other groups.
Second, the directional location metrics show that ASD fixations are the
\emph{hardest to predict}: any group predicting ASD scores lowest (e.g.\
\art\ra\asd\ $\text{AUC}=0.873$, \typ\ra\asd\ $=0.876$), whereas ASD as a
predictor is only slightly worse than the third group. Absolute agreements are
high across the board, all groups are drawn to the same strongly salient
regions, so the effect is a modulation on top of a large shared, stimulus-driven
scaffold rather than the presence or absence of alignment.

\begin{table}[tbp]
\centering\small
\caption{Spatial directed comparison (mean over 30 paintings). Higher is better
except KL. \art=Artists, \typ=Neurotypical, \asd=ASD.}
\label{tab:spatial}
\setlength{\tabcolsep}{3.5pt}
\begin{tabular}{lcccccc}
\toprule
pair & AUC$\uparrow$ & NSS$\uparrow$ & CC$\uparrow$ & SIM$\uparrow$ & KL$\downarrow$ & IG$\uparrow$\\
\midrule
\art\pto\typ & \textbf{0.908} & \textbf{2.094} & \textbf{0.959} & \textbf{0.856} & \textbf{0.092} & \textbf{0.980}\\
\typ\pto\art & 0.888 & 1.909 & \textbf{0.959} & \textbf{0.856} & 0.114 & 0.832\\
\asd\pto\typ & 0.902 & 2.035 & 0.937 & 0.822 & 0.156 & 0.866\\
\typ\pto\asd & 0.876 & 1.877 & 0.937 & 0.822 & 0.216 & 0.724\\
\asd\pto\art & 0.879 & 1.852 & 0.938 & 0.831 & 0.171 & 0.724\\
\art\pto\asd & 0.873 & 1.871 & 0.938 & 0.831 & 0.198 & 0.733\\
\bottomrule
\end{tabular}
\end{table}

The primary contrast (\asd\ra\art\ vs.\ \asd\ra\typ) is significant on the
directional metrics (AUC $d=-1.65$, NSS $d=-1.52$, IG $d=-0.89$; all
$p<10^{-4}$) but \emph{not} on the symmetric ones (CC $d=0.03$, $p=0.88$; SIM
$d=0.34$, $p=0.07$). The directional difference reflects that neurotypical
fixations are more predictable in general, not that ASD is selectively similar
to either group, symmetrically, ASD is equidistant from artists and
neurotypicals.

\subsection{Per-scanpath features: ASD explores widely but briefly}
\begin{figure}[tbp]
  \centering
  \includegraphics[width=\linewidth]{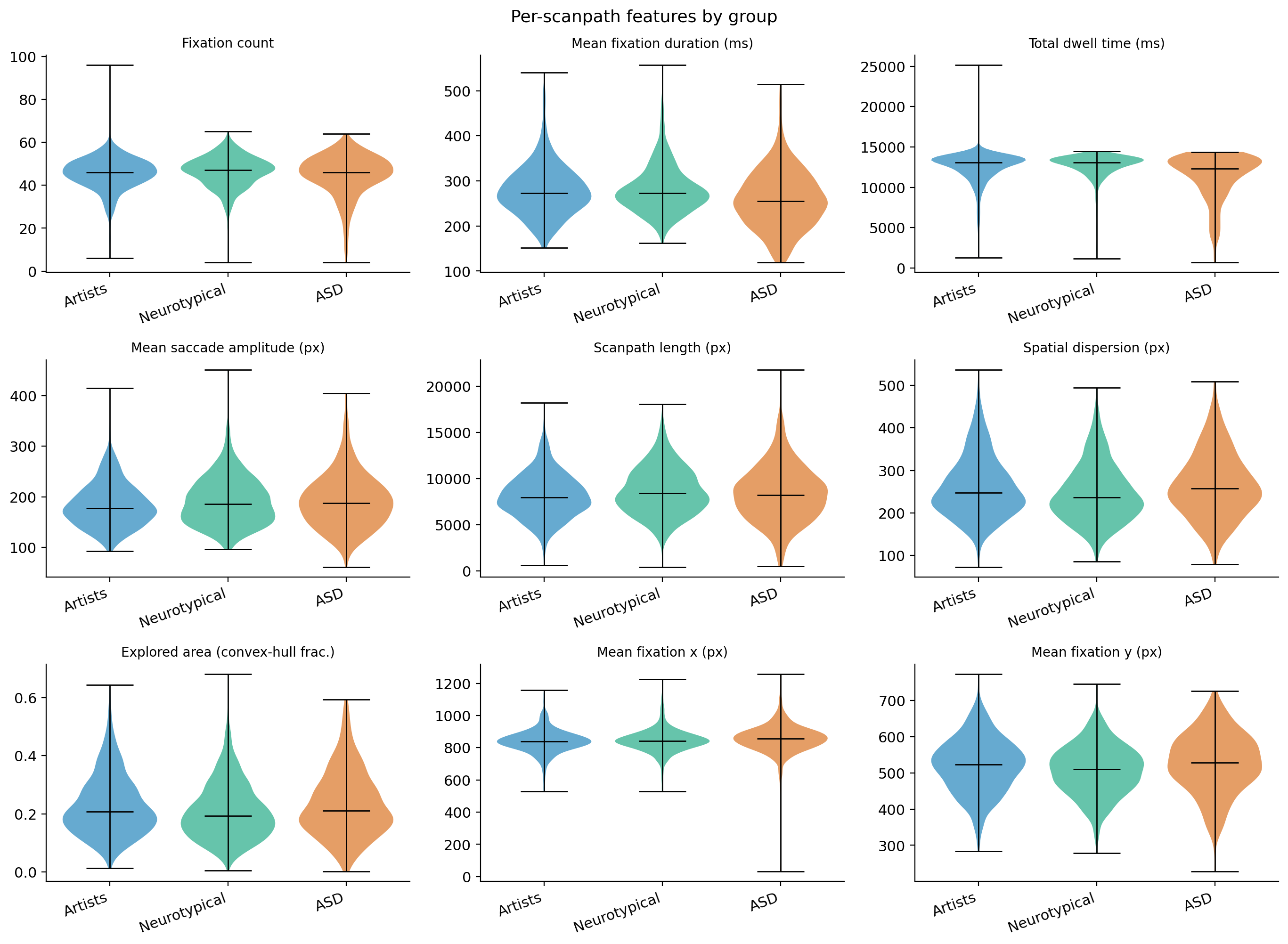}
  \caption{Per-scanpath feature distributions by group. ASD (orange) shows
  shorter fixations and lower total dwell, but dispersion and explored area
  comparable to artists (blue) and above neurotypicals (green).}
  \label{fig:feat}
\end{figure}

Table~\ref{tab:feat} reports per-scanpath features by group; all nine differ
across groups (Kruskal--Wallis, $p<0.03$; $p<10^{-3}$ for seven of nine).
ASD viewers make fixations that are \textbf{significantly shorter}
(258\,ms vs.\ 279/281\,ms; Holm $p<10^{-4}$) and accumulate \textbf{less total
dwell time}, despite a similar number of fixations. Spatially, ASD scanpaths are
\textbf{more dispersed} and cover a \textbf{larger explored area} than
neurotypical ones ($p<10^{-3}$), matching artists, and are the
\textbf{least centre-biased} (fixation centroid shifted right and downward,
$p<10^{-3}$). In short (Fig.~\ref{fig:feat}), ASD shares artists' wide spatial
exploration but couples it with a distinct temporal profile of shorter, less
dwelling fixations.

\begin{table}[tbp]
\centering\small
\caption{Per-scanpath features (mean). $H,p$: Kruskal--Wallis omnibus.
Dur.: fixation duration; Disp.: dispersion; Hull: explored-area fraction.}
\label{tab:feat}
\setlength{\tabcolsep}{4pt}
\begin{tabular}{lcccc}
\toprule
feature & \art & \typ & \asd & $p$\\
\midrule
Fixations                 & 45.3 & 46.2 & 44.6 & 0.02\\
Dur.\ mean (ms)           & 278.8 & 280.6 & \textbf{258.3} & $<10^{-4}$\\
Total dwell (ms)          & 12388 & 12680 & \textbf{11318} & $<10^{-4}$\\
Saccade amp.\ (px)        & 183.6 & 191.4 & 192.6 & 0.003\\
Scanpath length (px)      & 8148 & 8670 & 8373 & $<10^{-3}$\\
Dispersion (px)           & 261.3 & \textbf{246.7} & \textbf{265.5} & $<10^{-4}$\\
Explored area (hull)      & 0.23 & \textbf{0.21} & \textbf{0.23} & $10^{-4}$\\
Centroid $x$ (px)         & 839.6 & 843.9 & \textbf{856.0} & $10^{-4}$\\
Centroid $y$ (px)         & 520.7 & \textbf{507.3} & \textbf{526.5} & $<10^{-4}$\\
\bottomrule
\end{tabular}
\end{table}

\subsection{Temporal axis: ASD scanpaths are the most idiosyncratic}
\begin{figure*}[tbp]
  \centering
  \includegraphics[width=0.72\textwidth]{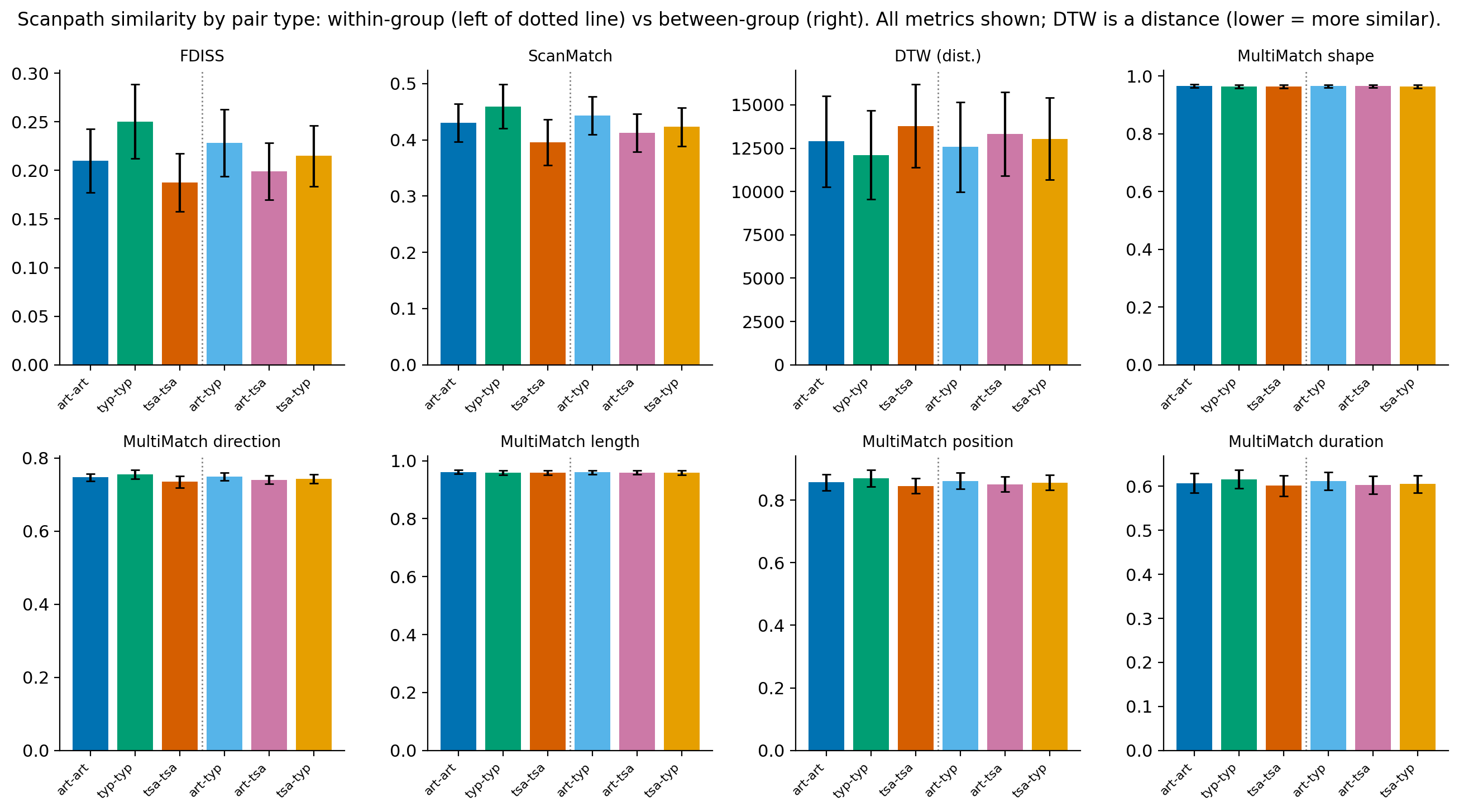}
  \caption{Scanpath similarity for all four measures and five MultiMatch
  sub-dimensions, within-group (left of dotted line) vs.\ between-group (right).
  On the discriminative measures (FDISS, ScanMatch, DTW, MultiMatch position):
  \texttt{typ-typ}$>$\texttt{art-art}$>$\texttt{tsa-tsa}, and \texttt{art-typ}
  is the most alignable between-group pair (DTW is a distance: lower is more
  similar). The MultiMatch shape, direction and length sub-dimensions saturate
  near ceiling for scanpaths of this length and are weakly discriminative.}
  \label{fig:scan}
\end{figure*}

\begin{figure*}[tbp]
  \centering
  \includegraphics[width=0.66\textwidth]{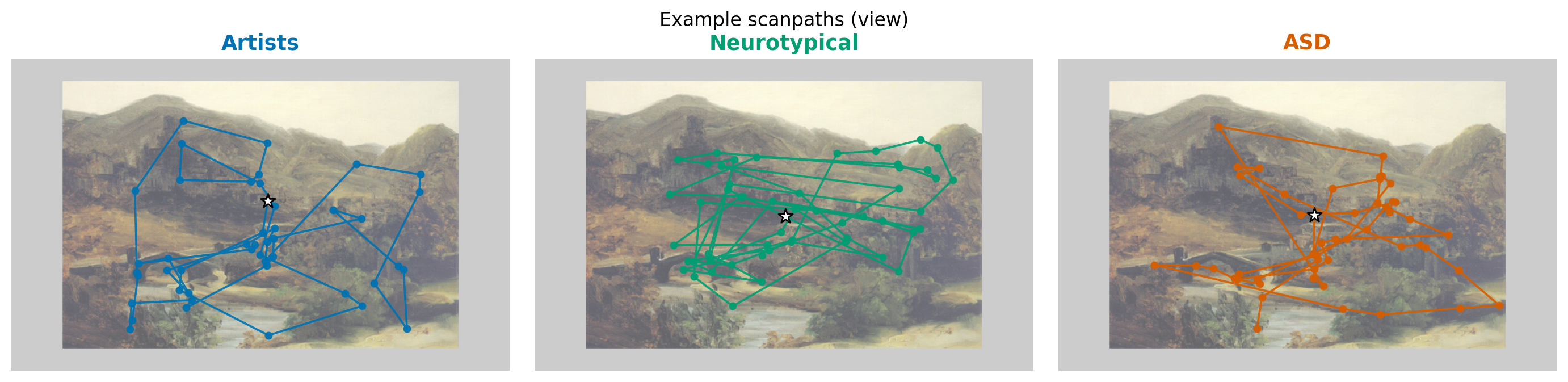}
  \caption{Representative scanpaths (one median-length participant per group) on
  an example painting. Star = first fixation; lines connect successive
  fixations. ASD trajectories are wider and less centred; neurotypical ones more
  compact.}
  \label{fig:scanex}
\end{figure*}

Table~\ref{tab:scan} reports within- and between-group scanpath similarity, and
the ordering is remarkably consistent across MultiMatch, ScanMatch, FDISS and
DTW (Fig.~\ref{fig:scan}). Figure~\ref{fig:scanex} illustrates the qualitative
difference with one representative scanpath per group. Three findings mirror and
extend the spatial axis.

\emph{Within-group cohesion.} Neurotypical scanpaths are the most
self-consistent, artists intermediate, and \textbf{ASD the least}
(\texttt{tsa-tsa} lowest on every metric: FDISS 0.188 vs.\ 0.210/0.250; ScanMatch
0.395 vs.\ 0.430/0.460; DTW largest distance). Autistic viewers are as different
from one another as they are from other groups, their ``group-specific''
pattern is better described as a set of heterogeneous individual styles than a
single shared one.

\emph{Between-group alignment.} \texttt{art--typ} is the most alignable
between-group pair, significantly exceeding both \texttt{art--tsa} (FDISS
$d=+2.32$) and \texttt{tsa--typ} (FDISS $d=+1.01$; all $p<10^{-3}$), reinforcing
the artist--neurotypical convergence seen spatially. The ordering is
stimulus-general rather than driven by a few paintings: by FDISS, \texttt{art--typ}
is the most alignable between-group pair on 25 of 30 paintings, and
\texttt{tsa--typ} exceeds \texttt{art--tsa} on 29 of 30 (spatially, \art--\typ\
has the highest CC on 26 of 30).

\emph{Not selectively artist-like.} The key temporal contrast
(\texttt{art--tsa} vs.\ \texttt{tsa--typ}) is significant on every metric, and in
the direction \emph{opposite} to an ``ASD-is-artist-like'' hypothesis: ASD
scanpaths are marginally \emph{more} similar to neurotypical than to artist ones
(FDISS $d=-1.99$, ScanMatch $d=-0.98$, DTW $d=+0.86$, MultiMatch-position
$d=-1.41$; all $p<0.002$).

\begin{table}[tbp]
\centering\small
\caption{Scanpath similarity per pair-type (mean over paintings). FDISS,
ScanMatch, MultiMatch-position: higher\,=\,more similar; DTW: lower\,=\,more
similar. Within-group cells shaded.}
\label{tab:scan}
\setlength{\tabcolsep}{4pt}
\begin{tabular}{lcccc}
\toprule
pair & FDISS$\uparrow$ & ScanM.$\uparrow$ & DTW$\downarrow$ & MM-pos$\uparrow$\\
\midrule
\rowcolor{black!6} art-art & 0.210 & 0.430 & 12888 & 0.857\\
\rowcolor{black!6} typ-typ & \textbf{0.250} & \textbf{0.460} & \textbf{12114} & \textbf{0.869}\\
\rowcolor{black!6} tsa-tsa & 0.188 & 0.395 & 13781 & 0.846\\
art-typ & 0.228 & 0.443 & 12569 & 0.862\\
art-tsa & 0.199 & 0.413 & 13318 & 0.851\\
tsa-typ & 0.215 & 0.423 & 13047 & 0.856\\
\bottomrule
\end{tabular}
\end{table}

\clearpage
\section{Stimulus Properties and Low-Level Determinants of Gaze}
Having established the group ordering in space and time, we now characterise the
stimuli themselves and the low-level and individual-level structure of gaze.
Table~\ref{tab:lowlevel} consolidates the group contrasts that recur below.

\begin{table}[tbp]
\centering\small
\caption{Consolidated group-level low-level differences (medians; Kruskal–Wallis
across groups). ASD is the least centre-biased, makes the smallest saccades,
lands off the painting most often, and is the least self-congruent group.}
\label{tab:lowlevel}
\setlength{\tabcolsep}{5pt}
\begin{tabular}{lcccc}
\toprule
measure & \art & \typ & \asd & KW $p$\\
\midrule
Central distance (norm.) & 0.41 & 0.39 & 0.42 & $<10^{-4}$\\
Saccade amplitude (°)    & 3.61 & 3.80 & 3.56 & $<10^{-10}$\\
Pupil diameter (mm)      & 4.46 & 4.93 & 4.90 & $<10^{-300}$\\
On-painting fraction     & 0.95 & 0.97 & 0.94 & ---\\
IOC (CC)                 & 0.67 & 0.74 & 0.61 & $<10^{-5}$\\
\bottomrule
\end{tabular}
\end{table}

\subsection{Stimulus image properties}
\begin{figure}[tbp]
  \centering
  \includegraphics[width=\linewidth]{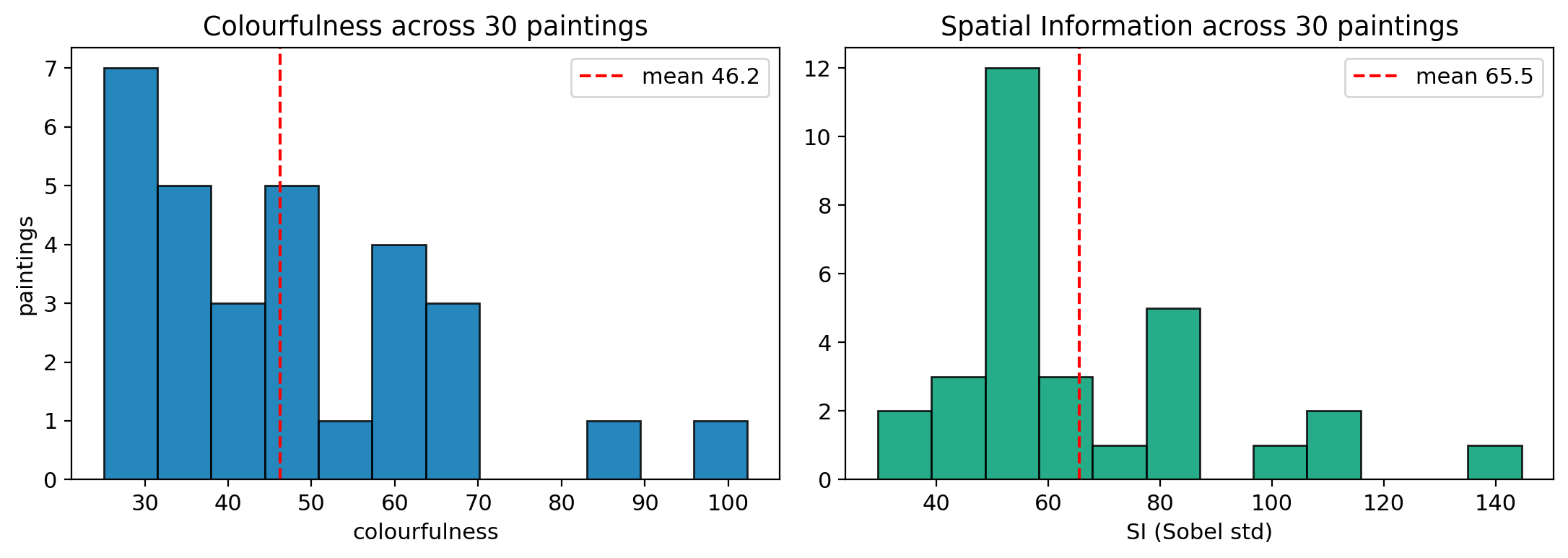}
  \caption{Univariate distributions of colourfulness and spatial information
  across the 30 paintings (grey borders excluded).}
  \label{fig:props-uni}
\end{figure}

\begin{figure}[tbp]
  \centering
  \includegraphics[width=0.82\linewidth]{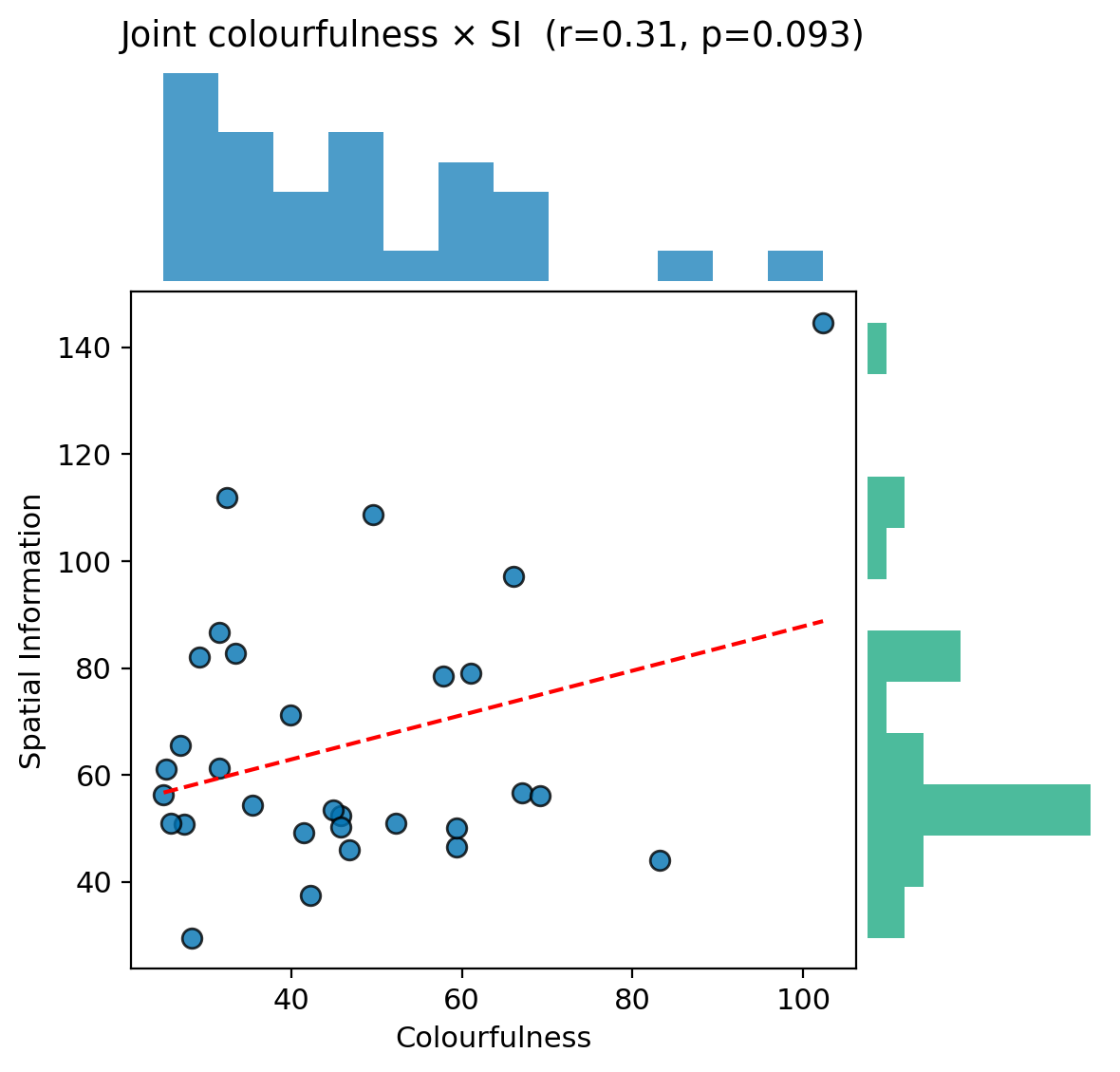}
  \caption{Joint colourfulness × SI with marginals; the two complexity axes are
  largely independent ($r=0.31$, n.s.).}
  \label{fig:props-joint}
\end{figure}

We first quantify the paintings themselves on content pixels only, having removed
the uniform grey letterbox (RGB 170,170,170) in which each image was displayed.
Per painting we compute \emph{colourfulness} (Hasler–Süsstrunk) and
\emph{spatial information} (SI; the standard deviation of the Sobel gradient of
luminance, ITU-T P.910). The set spans a wide range
(colourfulness $46.2\pm18.8$, range $25{-}102$; SI $65.5\pm25.0$, range
$30{-}145$; Fig.~\ref{fig:props-uni}, Table~\ref{tab:props}). The two properties
are only weakly and non-significantly correlated
($r=0.31$, $p=0.09$; Fig.~\ref{fig:props-joint}), so colour richness and
edge/detail density are largely independent axes of stimulus complexity here.
The palette is warm-dominated, with hue mass concentrated in reds–oranges–yellows,
mid-range saturation, and a broad brightness distribution
(Fig.~\ref{fig:hsv}; Hue$\times$Saturation density in Supplement Fig.~\ref{fig:s-huesat}).

\begin{figure*}[tbp]
  \centering
  \includegraphics[width=0.96\textwidth]{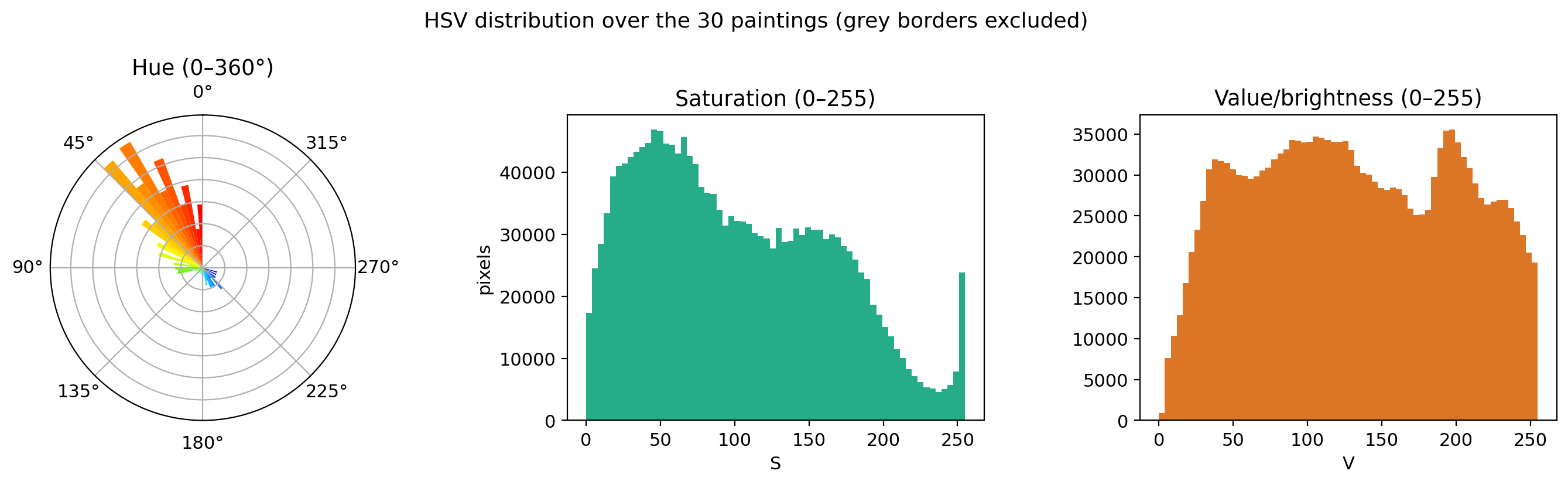}
  \caption{HSV distribution over all 30 paintings (content pixels only): warm-hue
  dominance, mid saturation, broad brightness.}
  \label{fig:hsv}
\end{figure*}

\begin{table}[tbp]
\centering\small
\caption{Painting properties (mean $\pm$ SD over 30 paintings; content pixels).}
\label{tab:props}
\begin{tabular}{lccccc}
\toprule
 & Colourf. & SI & Hue & Sat. & Val.\\
\midrule
mean & 46.2 & 65.5 & 48.4 & 101.0 & 131.2\\
SD   & 18.8 & 25.0 & 20.3 & 31.3 & 28.9\\
\bottomrule
\end{tabular}
\end{table}

\subsection{Saccadic dynamics}
\begin{figure*}[tbp]
  \centering
  \includegraphics[width=0.96\textwidth]{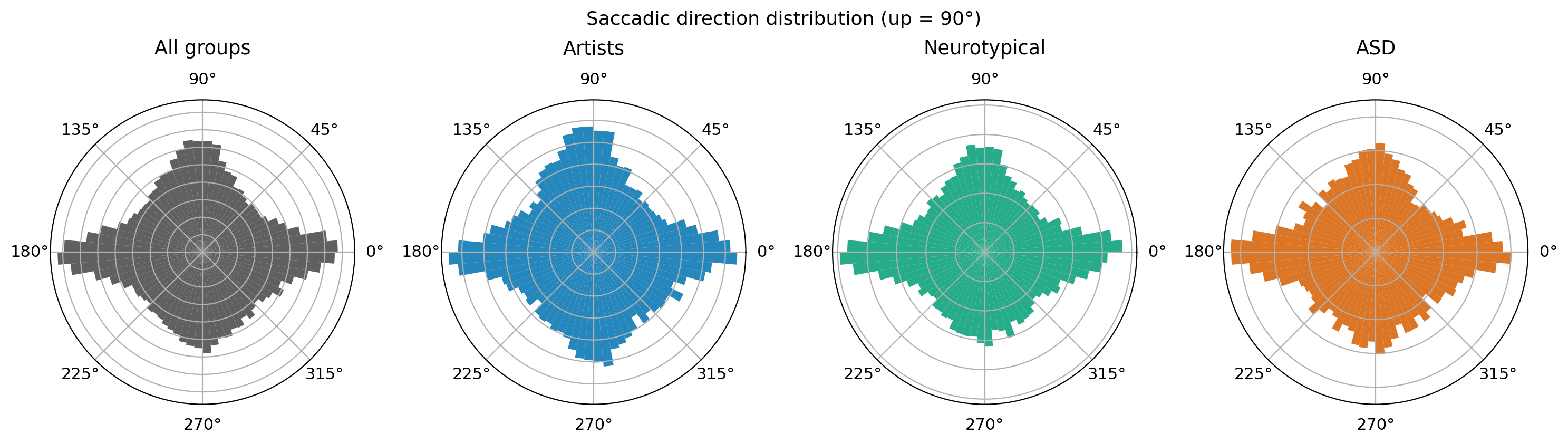}
  \caption{Saccadic direction (up $=90^\circ$): horizontal dominance in every
  group.}
  \label{fig:rose}
\end{figure*}

\begin{figure*}[tbp]
  \centering
  \includegraphics[width=0.96\textwidth]{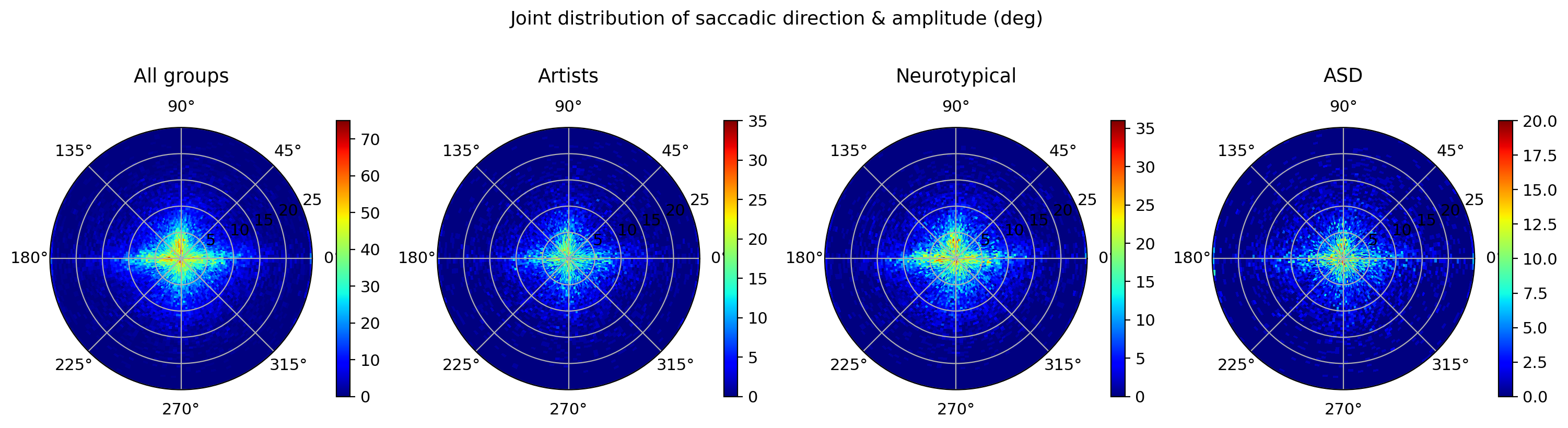}
  \caption{Joint distribution of saccadic direction and amplitude (degrees), per
  group and overall.}
  \label{fig:joint-sac}
\end{figure*}

\begin{figure}[tbp]
  \centering
  \includegraphics[width=\linewidth]{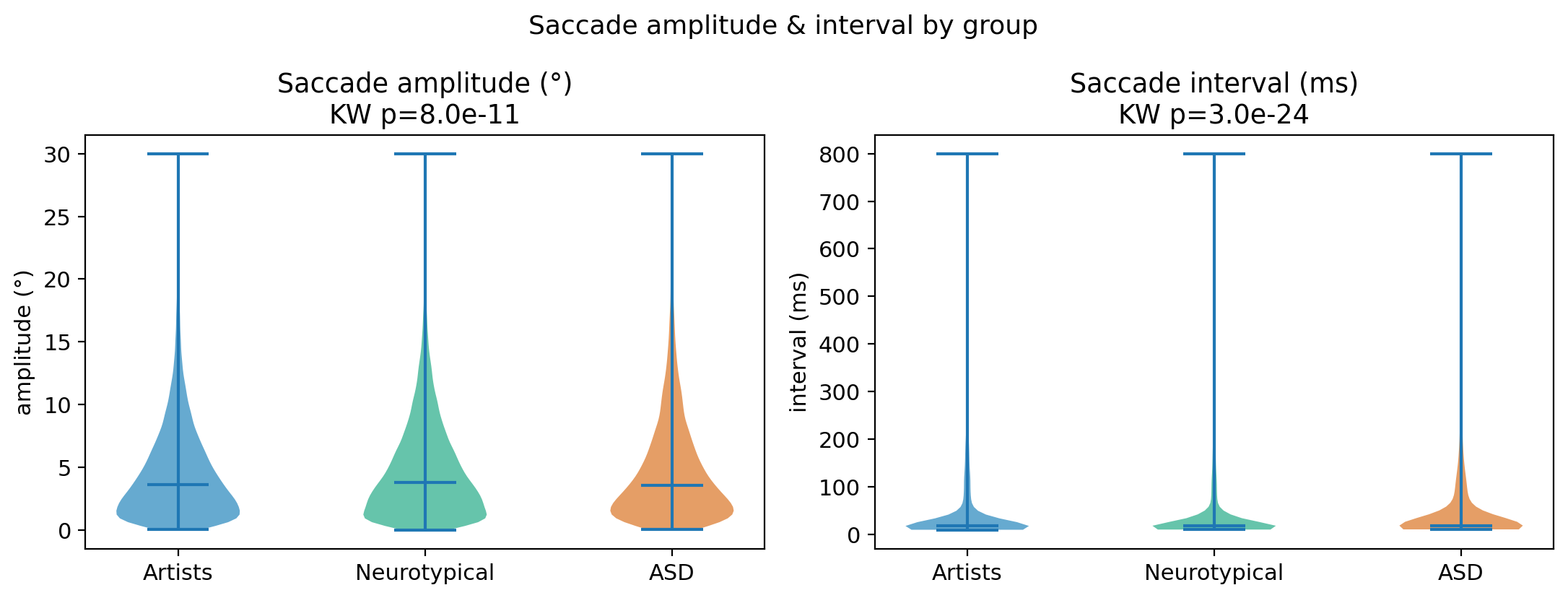}
  \caption{Saccade amplitude and inter-fixation interval by group.}
  \label{fig:sac-group}
\end{figure}

Saccades (transitions between consecutive fixations) are strongly anisotropic:
the direction distribution is dominated by the horizontal axis in all three
groups (Fig.~\ref{fig:rose}), and the joint direction × amplitude distribution
(Fig.~\ref{fig:joint-sac}) shows the characteristic horizontal "cross" with mass
concentrated at small amplitudes. Amplitudes differ by group
(median $3.8^\circ$ Typ, $3.6^\circ$ Art, $3.6^\circ$ ASD;
Kruskal–Wallis $p=8\times10^{-11}$; Fig.~\ref{fig:sac-group}): neurotypical
observers make the largest saccades, consistent with their broader, more
centre-anchored scanning, whereas ASD makes the smallest. The amplitude–interval
main sequence is shown in Supplement Fig.~\ref{fig:s-mainseq}.

\subsection{Spatial fixation distribution and central bias}
\begin{figure*}[tbp]
  \centering
  \includegraphics[width=0.96\textwidth]{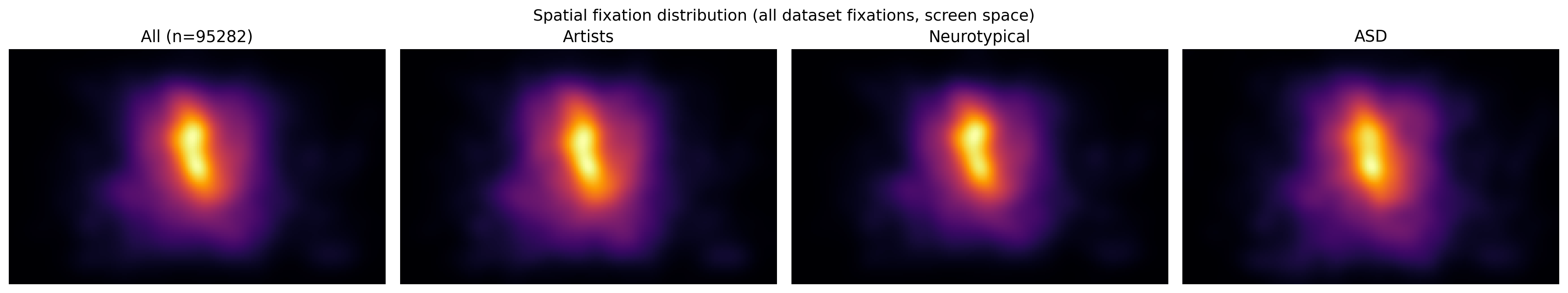}
  \caption{Spatial fixation density (all dataset fixations, screen space), per
  group and overall.}
  \label{fig:fixheat}
\end{figure*}

Pooling all $95{,}282$ fixations reveals a strong central bias in every group
(Fig.~\ref{fig:fixheat}), but its strength differs: neurotypical fixations are
the most centre-anchored and ASD the least (median normalised distance from
centre $0.39$ Typ $<0.41$ Art $<0.42$ ASD; Kruskal–Wallis $p<10^{-4}$; pairwise
Holm $p<10^{-4}$ for both ASD/Typ and Art/Typ; Fig.~\ref{fig:cbias}, radial
profile in Supplement Fig.~\ref{fig:s-radial}). ASD observers also land off the painting (on the
grey surround) more often than the others ($4.3\%$ vs.\ $3.5\%$ neurotypical).
The reduced central bias converges with the wider spatial dispersion reported in
Section~5.

\begin{figure}[tbp]
  \centering
  \includegraphics[width=0.86\linewidth]{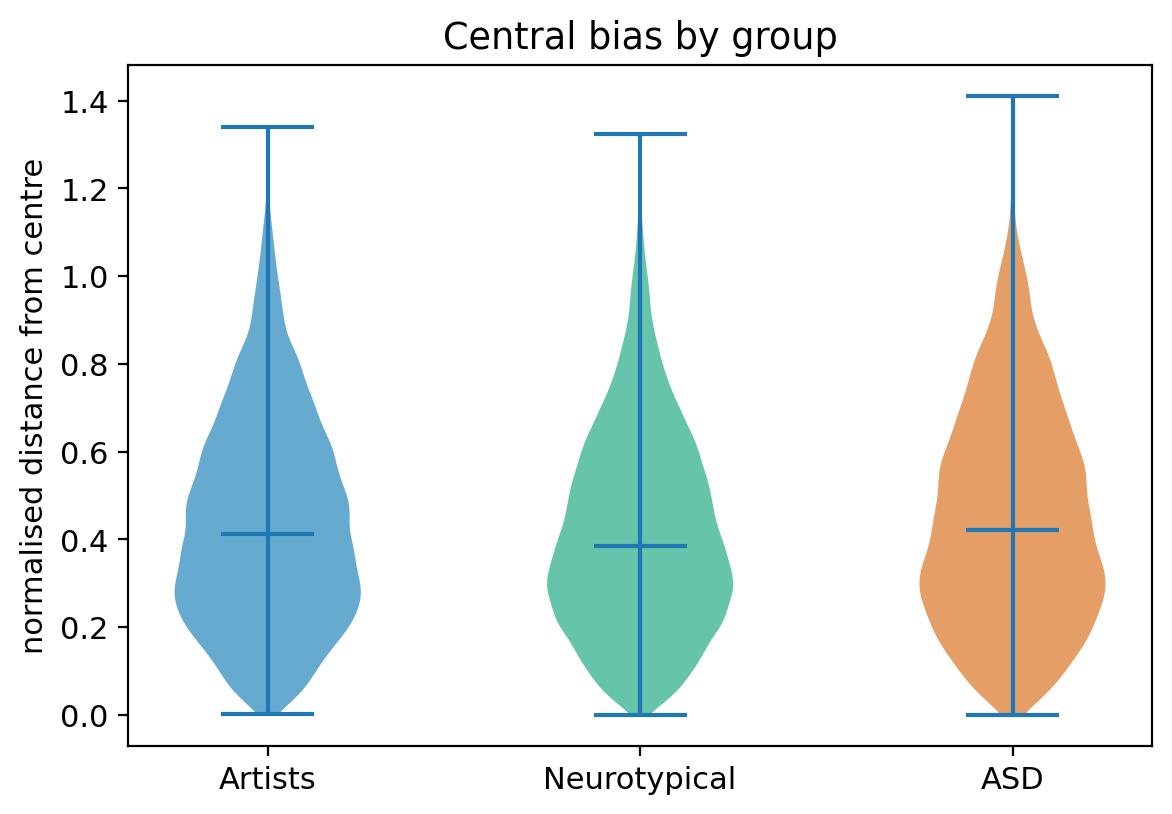}
  \caption{Central bias (normalised distance from screen centre) by group.}
  \label{fig:cbias}
\end{figure}

\subsection{Colour selection at fixation}
Sampling each fixation's painting pixel shows that gaze is not colour-neutral: in
all three groups, fixated pixels are significantly more \emph{saturated} (median
$103$ vs.\ available $92$; Mann–Whitney $p<10^{-140}$) and slightly
\emph{brighter} (median $133$ vs.\ $128$; $p<10^{-28}$) than the painting average
(Fig.~\ref{fig:coloravail}, per-hue preference in Supplement Fig.~\ref{fig:s-huepref}). The
preference for saturated, luminous regions is shared across groups, indicating a
common low-level colour driver of attention on top of the group-specific spatial
and temporal differences.

\begin{figure*}[tbp]
  \centering
  \includegraphics[width=0.96\textwidth]{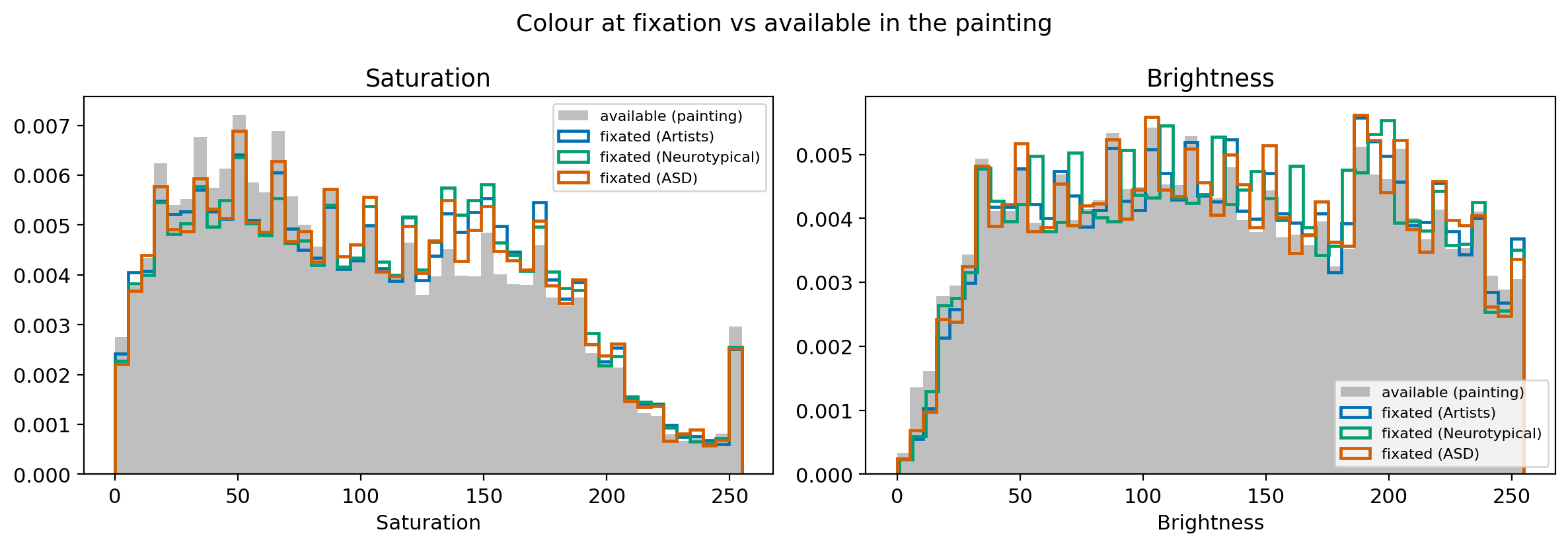}
  \caption{Colour at fixation vs.\ colour available in the painting; fixations
  shift toward higher saturation and brightness in every group.}
  \label{fig:coloravail}
\end{figure*}

\subsection{Pupillometry}
\begin{figure*}[tbp]
  \centering
  \includegraphics[width=0.96\textwidth]{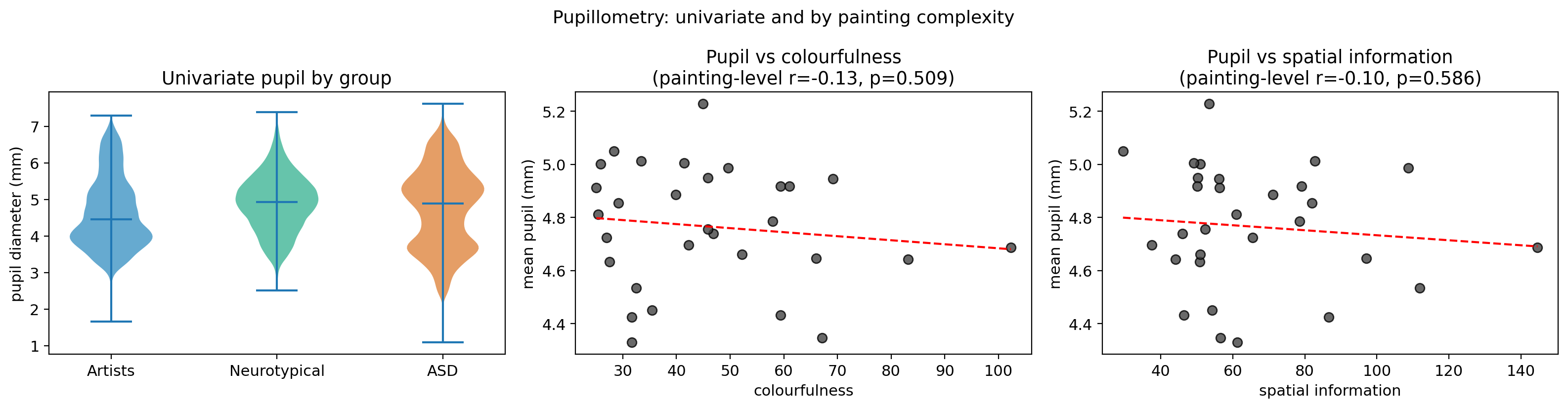}
  \caption{Pupillometry: univariate by group, and painting-level pupil vs.\
  colourfulness and spatial information.}
  \label{fig:pupil}
\end{figure*}

\begin{figure}[tbp]
  \centering
  \includegraphics[width=\linewidth]{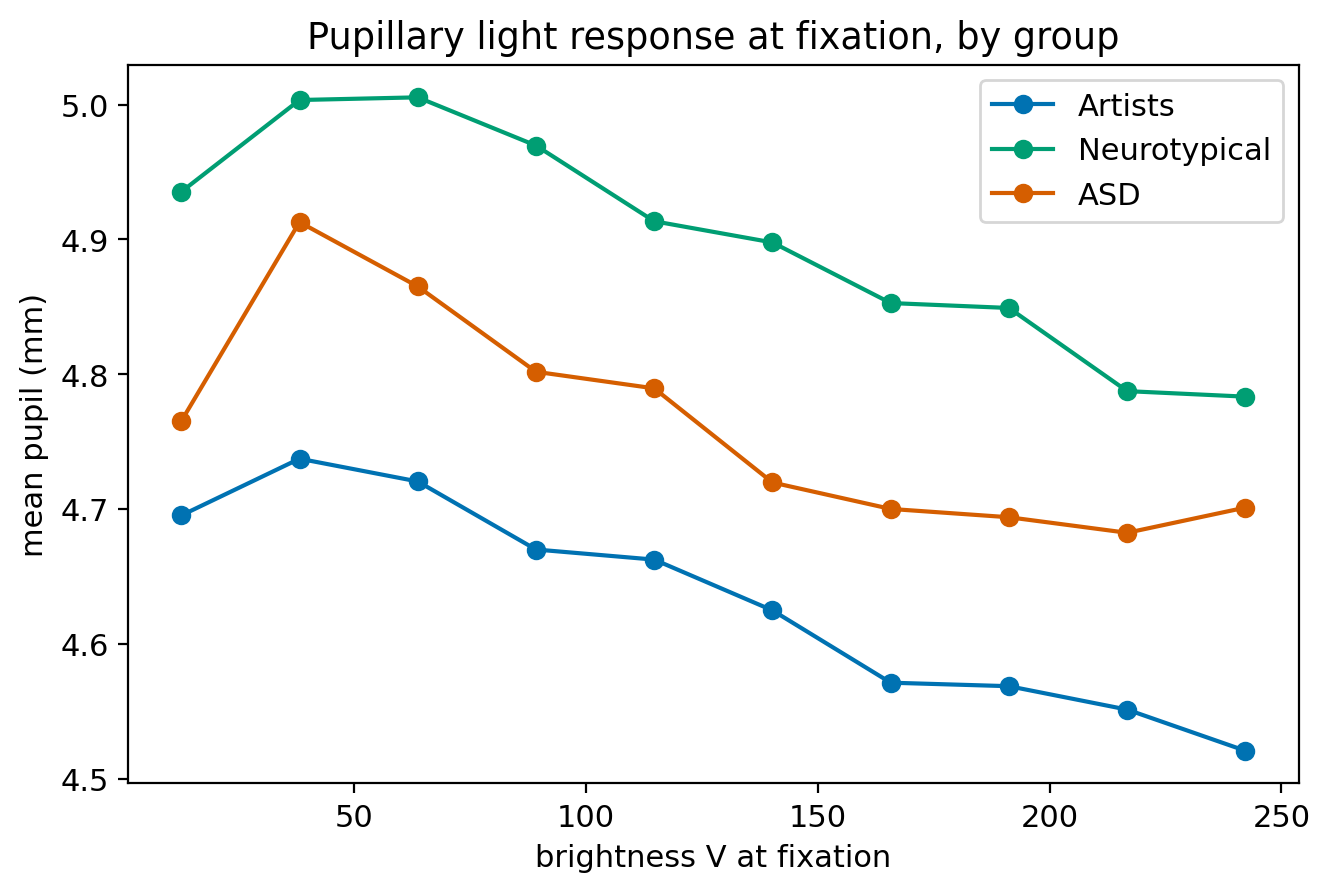}
  \caption{Pupillary light response at fixation (pupil vs.\ fixated brightness),
  by group.}
  \label{fig:pupil-light}
\end{figure}

Mean pupil diameter differs by group, with artists showing the smallest pupils
($4.46$\,mm) and neurotypical and ASD observers larger and similar
($4.93$/$4.90$\,mm; Fig.~\ref{fig:pupil}). Pupil size tracks the brightness at
the fixated location in the expected direction of the pupillary light response —
pupils constrict on brighter regions — in every group (Art $r=-0.07$, Typ
$r=-0.09$, ASD $r=-0.06$; all $p<10^{-16}$; Fig.~\ref{fig:pupil-light}), with the
response weakest in ASD. At the painting level, pupil size is only weakly related
to colourfulness and SI (Fig.~\ref{fig:pupil}, right), so the dominant modulator
of pupil here is local luminance rather than global stimulus complexity.

\subsection{Inter-observer congruency}
\begin{figure*}[tbp]
  \centering
  \includegraphics[width=0.96\textwidth]{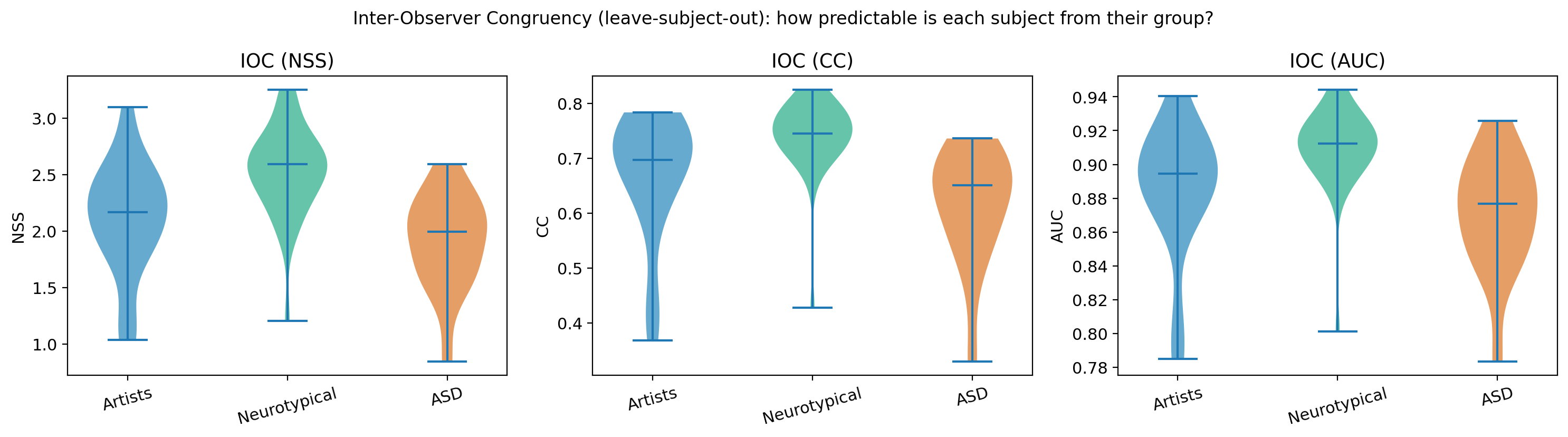}
  \caption{Inter-observer congruency (leave-subject-out) by group: ASD observers
  are the least predictable from their peers.}
  \label{fig:ioc}
\end{figure*}

Finally, a leave-one-subject-out analysis quantifies how predictable each
observer is from the rest of their own group: each subject's fixations are scored
(NSS, CC, AUC-Judd) against the density map of the remaining same-group members.
Congruency is ordered Neurotypical $>$ Artists $>$ ASD on every measure
(CC $0.74>0.67>0.61$; Kruskal–Wallis $p<10^{-5}$, $\varepsilon^2\approx0.34$;
Table~\ref{tab:ioc}, Fig.~\ref{fig:ioc}). Neurotypical observers are
significantly more congruent than both other groups, and ASD the least
(Holm-corrected CC: Typ$>$ASD $p<10^{-3}$, Typ$>$Art $p=0.004$, Art$>$ASD
$p=0.025$). This is an entirely independent confirmation, at the individual
level, of the within-group scanpath result in Section~5: autistic viewing is not
merely different from the other groups, it is the least internally consistent.

\begin{table}[tbp]
\centering\small
\caption{Inter-observer congruency (leave-subject-out), mean $\pm$ SD over
subjects. Higher = more predictable from one's own group.}
\label{tab:ioc}
\begin{tabular}{lccc}
\toprule
group & NSS & CC & AUC\\
\midrule
Artists      & $2.12\pm0.50$ & $0.67\pm0.11$ & $0.885\pm0.039$\\
Neurotypical & $\mathbf{2.53\pm0.40}$ & $\mathbf{0.74\pm0.07}$ & $\mathbf{0.910\pm0.025}$\\
ASD          & $1.91\pm0.43$ & $0.61\pm0.11$ & $0.871\pm0.035$\\
\bottomrule
\end{tabular}
\end{table}

\section{Discussion}
\paragraph{A convergent artist--neurotypical baseline.}
Across ten metrics and both axes, artists and neurotypicals are strikingly
similar: near-identical density maps and the most alignable scanpaths. Under
brief, untasked free-viewing, much of what both groups do is a shared,
conspicuity-driven exploration of the same compositional structure, and
expertise modulates gaze on top of that scaffold rather than replacing it. The
strong artist--neurotypical alignment should thus be read as evidence of shared
bottom-up looking under these conditions, not as proof that expertise leaves
gaze unchanged; a longer or explicitly analytic task would be expected to widen
the gap.

\paragraph{A spatial--temporal dissociation in ASD.}
The central new result is that autistic viewing cannot be summarised by a single
scalar of ``similarity.'' \emph{Spatially}, ASD gaze resembles artists: widely
dispersed, covering more of the canvas, and less centre-biased than neurotypical
gaze. \emph{Temporally}, it is distinct from both: fixations are shorter, dwell
is lower, and the ordered sequences are the least self-consistent of any group.
The same-\emph{where}, different-\emph{how} pattern is consistent with a
detail-focused, locally biased style \citep{happe2006weak,mottron2006enhanced}
and with gaze being tied more to low-level salience than to semantic structure
\citep{wang2015atypical}: rapid, broadly distributed sampling produces wide
spatial coverage but short, idiosyncratic temporal trajectories. Critically,
this is \emph{not} an exaggerated form of expert looking, on no metric is ASD
selectively artist-like.

\paragraph{Heterogeneity and personalised models.}
That \texttt{tsa-tsa} is the least cohesive cell indicates the ASD ``profile'' is
partly an aggregate of divergent individual strategies. This has a direct
methodological consequence: saliency and scanpath models for art are trained and
evaluated against neurotypical fixations \citep{kerkouri2022domain,
tliba2022selfsup,kerkouri2024avatt}, and our results imply that such models
predict artist gaze well but ASD gaze poorly, because viewer population is a
latent variable they ignore. Treating population identity, and, given the
heterogeneity, individual identity, as an explicit conditioning variable is a
concrete direction, especially for accessibility settings where the intended
viewer is by definition not neurotypical. The directed two-axis framework here
offers a ready way to audit such models: not only how well a model predicts
aggregate gaze, but \emph{whose} gaze, and whether it reproduces the temporal as
well as the spatial signature. The leave-subject-out inter-observer congruency
(Section~6.6) makes the heterogeneity concrete and quantitative: ASD observers
are the least predictable from their own group on every measure, independently
confirming the aggregate scanpath result at the level of individual subjects.

\paragraph{Low-level drivers beneath the group differences.}
The low-level analyses locate the group effects on a shared perceptual substrate.
All three groups fixate more saturated and brighter pixels than the painting
average, saccade predominantly along the horizontal, and show the pupillary light
response, so a common bottom-up machinery is engaged throughout. The group
differences ride on top of it: neurotypical gaze is the most centre-anchored and
makes the largest saccades, artists sit between, and ASD is the least
centre-biased, makes the smallest saccades, and most often leaves the painting
altogether. That colourfulness and spatial information are only weakly correlated
($r=0.31$) yet neither strongly modulates pupil size suggests that, under brief
free-viewing, local luminance rather than global stimulus complexity is the
dominant low-level driver, and that the group signature is carried by
\emph{how gaze is deployed} over that substrate rather than by differential
sensitivity to stimulus complexity.

\paragraph{A reusable two-axis auditing framework.}
Beyond the specific findings, the analysis itself is a contribution. Casting
group comparison as a \emph{directed} problem, one population's map or scanpath
used to predict another's gaze, turns any saliency or scanpath metric into a
measure of cross-group (dis)similarity, and pairing a spatial axis with a
temporal one exposes structure that either alone would miss. Here, a purely
spatial analysis would have reported that ASD ``explores like an artist,'' while
a purely temporal one would have missed the shared spatial breadth; only the two
together reveal the dissociation. The same framework generalises directly to
model auditing: a gaze model can be scored not only by aggregate accuracy but by
\emph{whose} gaze it reproduces and \emph{on which axis}, and its directional
(a)symmetries, for instance, that every group predicts ASD fixations better
than ASD predicts theirs, become first-class, testable quantities. Because the
pipeline applies one identical fixation-extraction rule to all groups and
reports ten complementary metrics with effect sizes and power, it offers a
transparent, reproducible template for characterising attentional differences in
other clinical or expert populations.

\section{Limitations}
Several limitations bound interpretation. The ASD sample is modest ($n=15$) and
the groups are not matched on age or sex, so group, age, sex and expertise are
partially confounded. Absolute cross-group similarities are high because pooled
maps over many participants are smooth and dominated by shared stimulus
structure; our conclusions are therefore comparative, about the ordering of
group pairings, rather than statements about absolute predictive quality.
Density maps pool participants and discard individual variability on the spatial
axis (the temporal axis, computed per participant, partially addresses this).
Fixation extraction uses one dispersion rule; while applied uniformly, different
thresholds would shift absolute counts (we report a robustness check in the
supplement). Finally, we do not yet include a computational bottom-up saliency
baseline or object/face/semantic annotations that would let us separate
stimulus-driven overlap from genuinely group-specific strategy.

\section{Conclusion \& Future Work}
We presented a directed, two-axis (spatial \emph{and} temporal),
metric-grounded comparison of free-viewing gaze on paintings across autistic,
artist and neurotypical observers. Artists and neurotypicals converge in both
space and time; ASD gaze is distinct on both axes, exhibits a spatial--temporal
dissociation (artist-like spatial breadth, a unique temporal signature), is the
most idiosyncratic group, and is not selectively artist-like. Future work should
scale the sample and match groups on demographics; move from pooled to
individual- and subgroup-level analyses to characterise ASD heterogeneity; add
bottom-up and semantic baselines to separate stimulus-driven from group-specific
gaze; and develop population- and person-conditioned models of aesthetic
attention, evaluated on both axes, for accessibility applications.

\section*{Declarations}
\footnotesize
\textbf{Ethics.} The study was conducted in accordance with the Declaration of
Helsinki. Participation was voluntary and informed consent was obtained from all
participants. Data were anonymised before analysis and no personally
identifiable information was collected or retained.

\smallskip\noindent\textbf{Data and code availability.} Analysis code, the
extracted fixation/scanpath tables, and all per-stimulus results are released at
\url{https://github.com/kmamine/TSA-Art-Typ-results}. Raw eye-tracking recordings
are available from the corresponding author on reasonable request, subject to the
participants' consent terms.

\smallskip\noindent\textbf{Author contributions (CRediT).}
M.A.K.: conceptualisation, methodology, software, formal analysis, writing --
original draft. D.S., R.J., O.L.: investigation, data curation. M.T., A.C.:
methodology, software, writing -- review \& editing. C.W., E.H.-D.,
S.M.-K.: investigation, resources, validation. N.A.-H.: conceptualisation,
supervision, funding acquisition, writing -- review \& editing.

\smallskip\noindent\textbf{Conflict of interest.} The authors declare no
competing interests.

\smallskip\noindent\textbf{Funding.} This work received no specific grant from
funding agencies in the public, commercial, or not-for-profit sectors.

\smallskip\noindent\textbf{Generative-AI usage disclosure.} An AI coding
assistant (Claude, Anthropic) was used under author supervision to help implement
the analysis pipeline, run the statistical computations, and draft and format the
manuscript. All methods, results, and interpretations were verified by the
authors, who take full responsibility for the content. No text or citation was
included without author verification.
\normalsize

\balance
\small
\bibliographystyle{plainnat}
\bibliography{references}

@article{bylinskii2018metrics,
author = {Bylinskii, Zoya and Judd, Tilke and Oliva, Aude and Torralba, Antonio and Durand, Fredo},
title = {What Do Different Evaluation Metrics Tell Us About Saliency Models?},
year = {2019},
issue_date = {March 2019},
publisher = {IEEE Computer Society},
address = {USA},
volume = {41},
number = {3},
issn = {0162-8828},
url = {https://doi.org/10.1109/TPAMI.2018.2815601},
doi = {10.1109/TPAMI.2018.2815601},
journal = {IEEE Trans. Pattern Anal. Mach. Intell.},
month = mar,
pages = {740–757},
numpages = {18}
}

@article{kummerer2015information,
  title={Information-theoretic model comparison unifies saliency metrics},
  author={K{\"u}mmerer, Matthias and Wallis, Thomas SA and Bethge, Matthias},
  journal={Proceedings of the National Academy of Sciences},
  volume={112},
  number={52},
  pages={16054--16059},
  year={2015},
  publisher={National Academy of Sciences}
}

@inproceedings{salvucci2000identifying,
  title={Identifying fixations and saccades in eye-tracking protocols},
  author={Salvucci, Dario D and Goldberg, Joseph H},
  booktitle={Proceedings of the 2000 symposium on Eye tracking research \& applications},
  pages={71--78},
  year={2000}
}

@article{dewhurst2012multimatch,
  title={It depends on how you look at it: Scanpath comparison in multiple dimensions with MultiMatch, a vector-based approach},
  author={Dewhurst, Richard and Nystr{\"o}m, Marcus and Jarodzka, Halszka and Foulsham, Tom and Johansson, Roger and Holmqvist, Kenneth},
  journal={Behavior research methods},
  volume={44},
  number={4},
  pages={1079--1100},
  year={2012},
  publisher={Springer}
}

@article{cristino2010scanmatch,
  title={ScanMatch: A novel method for comparing fixation sequences},
  author={Cristino, Filipe and Math{\^o}t, Sebastiaan and Theeuwes, Jan and Gilchrist, Iain D},
  journal={Behavior research methods},
  volume={42},
  number={3},
  pages={692--700},
  year={2010},
  publisher={Springer}
}

@misc{eyefeatures2024,
  title={eyefeatures: A {P}ython library for preprocessing, feature extraction and analysis of eye-movement data},
  author={Daudov, Vagiz and {contributors}}, year={2026},
  note={Python package, version 2.0.2},
  howpublished={\url{https://pypi.org/project/eyefeatures/}}
}

@phdthesis{kerkouri2024gaze,
  title={A Gaze into the Art World: Predicting Visual Attention Using Deep Learning},
  author={Kerkouri, Mohamed Amine},
  year={2024},
  school={Universit{\'e} d'Orl{\'e}ans}
}

@inproceedings{kerkouri2022domain,
  title={A domain adaptive deep learning solution for scanpath prediction of paintings},
  author={Kerkouri, Mohamed Amine and Tliba, Marouane and Chetouani, Aladine and Bruno, Alessandro},
  booktitle={Proceedings of the 19th International Conference on Content-Based Multimedia Indexing},
  pages={57--63},
  year={2022}
}

@inproceedings{kerkouri2024avatt,
  title={AVAtt: Art Visual Attention dataset for diverse painting styles},
  author={Kerkouri, Mohamed Amine and Tliba, Marouane and Chetouani, Aladine and Bruno, Alessandro},
  booktitle={Proceedings of the 2024 Symposium on Eye Tracking Research and Applications},
  pages={1--3},
  year={2024}
}

@inproceedings{tliba2022selfsup,
  title={Self supervised scanpath prediction framework for painting images},
  author={Tliba, Marouane and Kerkouri, Mohamed Amine and Chetouani, Aladine and Bruno, Alessandro},
  booktitle={Proceedings of the IEEE/CVF Conference on Computer Vision and Pattern Recognition},
  pages={1539--1548},
  year={2022}
}

@inproceedings{kerkouri2026fdiss,
  title={Closing the Foveal Gap: Perceptually Grounded Scanpath Comparison with Disc IoU},
  author={Kerkouri, Mohamed Amine and Tliba, Marouane and Sellam, Zakaria and Distante, Cosimo and Bruno, Alessandro and Chetouani, Aladine},
  booktitle={Proceedings of the 2026 Symposium on Eye Tracking Research and Applications},
  pages={1--3},
  year={2026}
}

@inproceedings{kerkouri2026semantic,
  title={What They Saw, Not Just Where They Looked: Semantic Scanpath Similarity via VLMs and NLP metrics},
  author={Kerkouri, Mohamed Amine and Tliba, Marouane and Wang, Bin and Chetouani, Aladine and Bagci, Ulas and Bruno, Alessandro},
  booktitle={Proceedings of the 2026 Symposium on Eye Tracking Research and Applications},
  pages={1--7},
  year={2026}
}

@article{vogt2007expertise,
  title={Expertise in pictorial perception: Eye-movement patterns and visual memory in artists and laymen},
  author={Vogt, Stine and Magnussen, Svein},
  journal={Perception},
  volume={36},
  number={1},
  pages={91--100},
  year={2007},
  publisher={SAGE Publications Sage UK: London, England}
}

@article{francuz2018eye,
  title={Eye movement correlates of expertise in visual arts},
  author={Francuz, Piotr and Zaniewski, Iwo and Augustynowicz, Pawe{\l} and Kopi{\'s}, Natalia and Jankowski, Tomasz},
  journal={Frontiers in human neuroscience},
  volume={12},
  pages={87},
  year={2018},
  publisher={Frontiers Media SA}
}

@article{happe2006weak,
  title={The Weak Coherence Account: Detail-focused Cognitive Style in Autism Spectrum Disorders: Happ{\'e} and Frith},
  author={Happ{\'e}, Francesca and Frith, Uta},
  journal={Journal of autism and developmental disorders},
  volume={36},
  number={1},
  pages={5--25},
  year={2006},
  publisher={Springer}
}

@article{mottron2006enhanced,
  title={Enhanced Perceptual Functioning in Autism: An Update, and Eight Principles of Autistic Perception: Mottron, Dawson, Souli{\`e}res, Hubert, and Burack},
  author={Mottron, Laurent and Dawson, Michelle and Souli{\`e}res, Isabelle and Hubert, Benedicte and Burack, Jake},
  journal={Journal of autism and developmental disorders},
  volume={36},
  number={1},
  pages={27--43},
  year={2006},
  publisher={Springer}
}

@article{wang2015atypical,
  title={Atypical visual saliency in autism spectrum disorder quantified through model-based eye tracking},
  author={Wang, Shuo and Jiang, Ming and Duchesne, Xavier Morin and Laugeson, Elizabeth A and Kennedy, Daniel P and Adolphs, Ralph and Zhao, Qi},
  journal={Neuron},
  volume={88},
  number={3},
  pages={604--616},
  year={2015},
  publisher={Elsevier}
}

@article{pelphrey2002visual,
  title={Visual scanning of faces in autism},
  author={Pelphrey, Kevin A and Sasson, Noah J and Reznick, J Steven and Paul, Gregory and Goldman, Barbara D and Piven, Joseph},
  journal={Journal of autism and developmental disorders},
  volume={32},
  number={4},
  pages={249--261},
  year={2002},
  publisher={Springer}
}

@article{hedley2012using,
  title={Using Eye Movements as an Index of Implicit Face Recognition in A utism S pectrum D isorder},
  author={Hedley, Darren and Young, Robyn and Brewer, Neil},
  journal={Autism Research},
  volume={5},
  number={5},
  pages={363--379},
  year={2012},
  publisher={Wiley Online Library}
}

@article{ricou2025invariant,
  title={Invariant response to faces in ASD: Unexpected trajectory of oculo-pupillometric biomarkers from childhood to adulthood},
  author={Ricou, Camille and Mofid, Yassine and Roch{\'e}, Laetitia and Bufo, Maria Rosa and Houy-Durand, Emmanuelle and Malvy, Jo{\"e}lle and Lemaire, Mathieu and Elian, Jean-Claude and Martineau, Jo{\"e}lle and Bonnet-Brilhault, Fr{\'e}d{\'e}rique and others},
  journal={Brain Research},
  pages={150070},
  year={2025},
  publisher={Elsevier}
}

@article{harrison2021icd,
  title={ICD-11: an international classification of diseases for the twenty-first century},
  author={Harrison, James E and Weber, Stefanie and Jakob, Robert and Chute, Christopher G},
  journal={BMC medical informatics and decision making},
  volume={21},
  number={Suppl 6},
  pages={206},
  year={2021},
  publisher={Springer}
}

\clearpage
\onecolumn
\normalsize
\appendix
\renewcommand{\thesection}{S\arabic{section}}
\renewcommand{\thetable}{S\arabic{table}}
\renewcommand{\thefigure}{S\arabic{figure}}
\setcounter{section}{0}\setcounter{table}{0}\setcounter{figure}{0}

\begin{center}
{\Large\bfseries Supplementary Material}\\[0.3em]
{\normalsize Divergent Gaze Patterns in Artistic Viewing: Spatial \emph{and}
Temporal Signatures}
\end{center}
\vspace{0.5em}

This supplement provides the full statistical tables underlying the main results
(the main text reports condensed versions), an additional qualitative figure,
and the exact methodological parameters and software versions needed to
reproduce the analysis. All tables are regenerated by the released code from
the raw SMI exports.

\section{Fixation extraction and dataset details}
Fixations were extracted uniformly for all groups from the median-smoothed,
averaged binocular point of regard using dispersion-threshold identification
(IDT) \citep{salvucci2000identifying}. Parameters:
median-filter window $=15$ samples ($\approx$30\,ms at 500\,Hz);
down-sampling factor $=3$ (500\,\pto\,$\approx$167\,Hz) for tractable IDT;
minimum fixation duration $=100$\,ms; maximum duration $=1500$\,ms; maximum
dispersion $=90$\,px ($\approx 2.3^{\circ}$ at 39\,px/deg). Scanpaths with fewer
than four fixations were discarded. The final set contains 95{,}282 fixations in
2{,}091 scanpaths (Artists 712, Neurotypical 940, ASD 439), mean 45.6 fixations
per scanpath and mean fixation duration $\approx$200\,ms. The vendor per-sample
event labels were not used: on this recording they label $\sim$47\% of samples
as saccade, which fragments fixations and, because the fragmentation depends on
per-participant tracking quality, would confound group comparison.

Spatial density maps were built on a $210\times131$ grid (screen down-scaled
$8\times$), with a Gaussian kernel $\sigma=40$\,px ($\approx1^{\circ}$); the
Information-Gain baseline is a centred Gaussian prior. Scanpath metrics used a
foveal radius of $1^{\circ}$ (39\,px) for FDISS and a $12\times8$ ScanMatch grid.

\section{Spatial axis: full ASD selectivity contrast}
Table~\ref{tab:s-spatial} gives the paired contrast \asd\ra\art\ vs.\
\asd\ra\typ\ over the 30 paintings for all six metrics. The directional metrics
(AUC, NSS, IG) differ significantly (ASD predicts neurotypical fixations better
than artist fixations), but the symmetric distribution metrics (CC, SIM) do not:
ASD is spatially equidistant from the two groups.

\begin{table}[htbp]
\centering\small
\caption{Spatial primary contrast \asd\ra\art\ (\emph{a}) vs.\ \asd\ra\typ\
(\emph{b}), paired over 30 paintings.}
\label{tab:s-spatial}
\begin{tabular}{lccccccc}
\toprule
metric & mean$_a$ & mean$_b$ & diff & $t$ & $p$ & Cohen's $d$ & power\\
\midrule
AUC      & 0.879 & 0.902 & $-0.023$ & $-9.06$ & $<10^{-4}$ & $-1.65$ (L) & 1.00\\
NSS      & 1.853 & 2.035 & $-0.183$ & $-8.30$ & $<10^{-4}$ & $-1.52$ (L) & 1.00\\
CC       & 0.938 & 0.937 & $+0.001$ & $0.15$  & $0.884$    & $+0.03$ (T) & 0.05\\
SIM      & 0.831 & 0.822 & $+0.009$ & $1.87$  & $0.072$    & $+0.34$ (S) & 0.44\\
KL       & 0.171 & 0.156 & $+0.015$ & $1.28$  & $0.211$    & $+0.23$ (S) & 0.24\\
InfoGain & 0.724 & 0.866 & $-0.141$ & $-4.89$ & $<10^{-4}$ & $-0.89$ (L) & 1.00\\
\bottomrule
\end{tabular}
\end{table}

\section{Temporal axis: full similarity table}
Table~\ref{tab:s-scan} reports all four measures and the five MultiMatch
sub-dimensions for every pair type (mean over paintings; standard deviations are
in the released CSV). The MultiMatch shape, length and direction sub-dimensions
saturate near ceiling for the $\sim$46-fixation scanpaths here and are therefore
weakly discriminative; FDISS, ScanMatch, DTW and MultiMatch-position carry the
group structure.

\begin{table}[htbp]
\centering\small
\caption{Scanpath similarity per pair type (mean over 30 paintings). FDISS,
ScanMatch (SM), and all MultiMatch (MM) dimensions: higher = more similar;
DTW: lower = more similar. Within-group rows shaded.}
\label{tab:s-scan}
\setlength{\tabcolsep}{4.5pt}
\begin{tabular}{lccccccccc}
\toprule
pair & FDISS & SM & DTW & MM-shape & MM-dir & MM-len & MM-pos & MM-dur\\
\midrule
\rowcolor{black!6} art-art & 0.210 & 0.430 & 12888 & 0.966 & 0.747 & 0.961 & 0.857 & 0.607\\
\rowcolor{black!6} typ-typ & 0.250 & 0.460 & 12114 & 0.965 & 0.755 & 0.959 & 0.869 & 0.616\\
\rowcolor{black!6} tsa-tsa & 0.188 & 0.395 & 13781 & 0.964 & 0.735 & 0.959 & 0.846 & 0.601\\
art-typ & 0.228 & 0.443 & 12569 & 0.965 & 0.750 & 0.960 & 0.862 & 0.612\\
art-tsa & 0.199 & 0.413 & 13318 & 0.965 & 0.740 & 0.960 & 0.851 & 0.603\\
tsa-typ & 0.215 & 0.423 & 13047 & 0.964 & 0.743 & 0.959 & 0.856 & 0.605\\
\bottomrule
\end{tabular}
\end{table}

Table~\ref{tab:s-scancontrast} gives the temporal primary contrast
(\texttt{art-tsa} vs.\ \texttt{tsa-typ}); every metric is significant and in the
direction of ASD being closer to neurotypical than to artist. Table~%
\ref{tab:s-between} confirms the \texttt{art-typ} pair is the most alignable,
significantly exceeding both ASD-involving between-group pairs.

\begin{table}[htbp]
\centering\small
\caption{Temporal primary contrast \texttt{art-tsa} (\emph{a}) vs.\
\texttt{tsa-typ} (\emph{b}), paired over 30 paintings.}
\label{tab:s-scancontrast}
\begin{tabular}{lcccccc}
\toprule
metric & mean$_a$ & mean$_b$ & $t$ & $p$ & Cohen's $d$ & power\\
\midrule
FDISS        & 0.199 & 0.215 & $-10.92$ & $<10^{-4}$ & $-1.99$ (L) & 1.00\\
ScanMatch    & 0.413 & 0.423 & $-5.35$  & $<10^{-4}$ & $-0.98$ (L) & 1.00\\
DTW          & 13318 & 13047 & $4.71$   & $10^{-4}$  & $+0.86$ (L) & 1.00\\
MM-shape     & 0.965 & 0.964 & $3.47$   & $0.002$    & $+0.63$ (M) & 0.92\\
MM-direction & 0.740 & 0.743 & $-3.80$  & $<10^{-3}$ & $-0.69$ (M) & 0.96\\
MM-length    & 0.960 & 0.959 & $3.99$   & $<10^{-3}$ & $+0.73$ (M) & 0.97\\
MM-position  & 0.851 & 0.856 & $-7.73$  & $<10^{-4}$ & $-1.41$ (L) & 1.00\\
MM-duration  & 0.603 & 0.605 & $-1.63$  & $0.115$    & $-0.30$ (S) & 0.35\\
\bottomrule
\end{tabular}
\end{table}

\begin{table}[htbp]
\centering\small
\caption{Between-group ordering: \texttt{art-typ} vs.\ the two ASD-involving
between-group pairs (paired over paintings; Cohen's $d$, $p$).}
\label{tab:s-between}
\begin{tabular}{lcccc}
\toprule
 & \multicolumn{2}{c}{vs.\ \texttt{art-tsa}} & \multicolumn{2}{c}{vs.\ \texttt{tsa-typ}}\\
\cmidrule(lr){2-3}\cmidrule(lr){4-5}
metric & $d$ & $p$ & $d$ & $p$\\
\midrule
FDISS        & $+2.32$ & $<10^{-4}$ & $+1.01$ & $<10^{-4}$\\
ScanMatch    & $+1.72$ & $<10^{-4}$ & $+1.21$ & $<10^{-4}$\\
DTW          & $-1.11$ & $<10^{-4}$ & $-0.74$ & $<10^{-3}$\\
MM-direction & $+1.27$ & $<10^{-4}$ & $+0.80$ & $10^{-4}$\\
MM-position  & $+1.43$ & $<10^{-4}$ & $+0.80$ & $10^{-4}$\\
MM-duration  & $+1.20$ & $<10^{-4}$ & $+0.71$ & $<10^{-3}$\\
\bottomrule
\end{tabular}
\end{table}

\section{Per-scanpath features: pairwise ASD contrasts}
Table~\ref{tab:s-feat} reports Holm-corrected Mann--Whitney contrasts of ASD vs.\
the other two groups (rank-biserial correlation, rbc; positive = first group
higher). ASD has shorter fixations and less dwell than both groups, and greater
dispersion, larger explored area and a less centred fixation centroid than
neurotypicals.

\begin{table}[htbp]
\centering\small
\caption{Feature contrasts involving ASD (Mann--Whitney, Holm-corrected).
rbc: rank-biserial effect size.}
\label{tab:s-feat}
\begin{tabular}{llccc}
\toprule
feature & contrast & med$_1$ & med$_2$ & rbc ($p$)\\
\midrule
Dur.\ mean      & art\,/\,tsa & 272.8 & 254.5 & $-0.19\ (<10^{-4})$\\
                & tsa\,/\,typ & 254.5 & 272.7 & $+0.21\ (<10^{-4})$\\
Total dwell     & art\,/\,tsa & 13053 & 12338 & $-0.20\ (<10^{-4})$\\
                & tsa\,/\,typ & 12338 & 13074 & $+0.26\ (<10^{-4})$\\
Dispersion      & tsa\,/\,typ & 257.2 & 235.8 & $-0.15\ (<10^{-3})$\\
Explored area   & tsa\,/\,typ & 0.211 & 0.192 & $-0.12\ (<10^{-3})$\\
Centroid $x$    & art\,/\,tsa & 838.9 & 856.5 & $+0.14\ (10^{-4})$\\
Centroid $y$    & tsa\,/\,typ & 527.9 & 510.0 & $-0.16\ (<10^{-4})$\\
Saccade amp.    & art\,/\,tsa & 177.7 & 187.3 & $+0.09\ (0.02)$\\
\bottomrule
\end{tabular}
\end{table}

\section{Robustness to the fixation-extraction threshold}
Because IDT requires a dispersion threshold, we re-ran the full extraction at a
stricter ($60$\,px) and a looser ($120$\,px) value, bracketing the $90$\,px used
in the main text, and recomputed the headline quantities
(Table~\ref{tab:s-robust}). Absolute fixation counts and durations shift with the
threshold, as expected, but every conclusion is preserved at all three settings:
ASD has the shortest mean fixation duration; within-group cohesion is always
ordered \texttt{typ-typ}$>$\texttt{art-art}$>$\texttt{tsa-tsa}; and the ASD
selectivity contrast \texttt{art-tsa}$<$\texttt{tsa-typ} holds with a large
effect ($d\!\le\!-1.95$, $p<10^{-10}$). The headline findings are therefore not
artefacts of a particular threshold choice.

\begin{table}[htbp]
\centering\small
\caption{Robustness of the main conclusions to the IDT dispersion threshold
(FDISS for cohesion/contrast). Med.\ fixation duration in ms; cohesion is mean
within-group FDISS.}
\label{tab:s-robust}
\setlength{\tabcolsep}{4pt}
\begin{tabular}{lccc}
\toprule
quantity & $60$\,px & $90$\,px (main) & $120$\,px\\
\midrule
Fixations / scanpath          & 50.1 & 45.6 & 41.8\\
Dur.\ art / typ / tsa (ms)    & 232/226/\textbf{214} & 279/281/\textbf{258} & 313/308/\textbf{288}\\
Cohesion art / typ / tsa      & .22/.26/\textbf{.19} & .21/.25/\textbf{.19} & .20/.24/\textbf{.18}\\
\texttt{art-tsa} / \texttt{tsa-typ} & .204/.222 & .199/.215 & .191/.207\\
Contrast $d$ ($p$)            & $-2.08\ (10^{-12})$ & $-1.99\ (10^{-11})$ & $-1.95\ (10^{-11})$\\
\bottomrule
\end{tabular}
\end{table}

\section{Additional low-level figures}
This section collects the supporting figures referenced from
Section~6 (stimulus properties and low-level determinants of gaze).

\begin{figure}[htbp]
  \centering
  \includegraphics[width=0.58\linewidth]{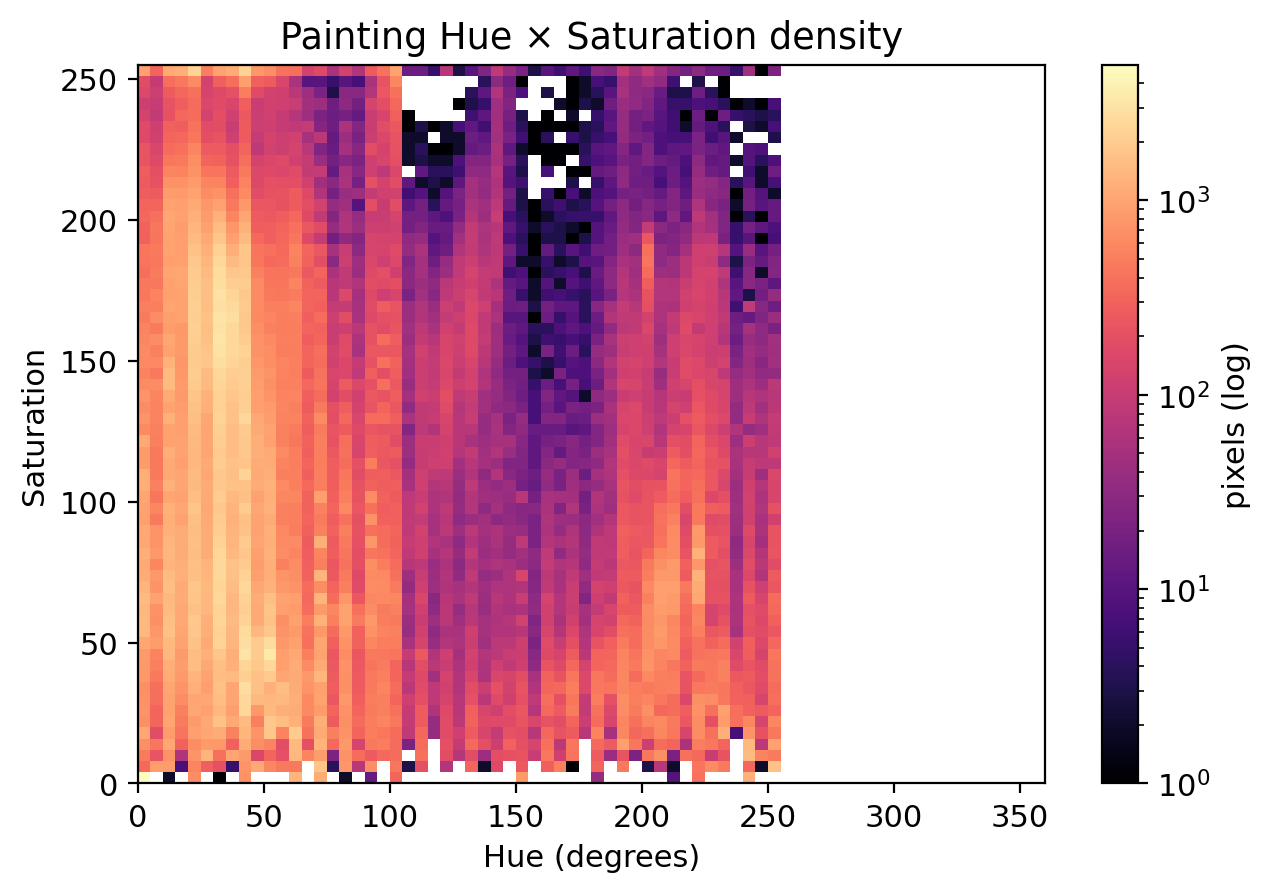}
  \caption{Painting Hue $\times$ Saturation density (content pixels, log scale):
  warm hues at mid-to-high saturation dominate.}
  \label{fig:s-huesat}
\end{figure}

\begin{figure}[htbp]
  \centering
  \includegraphics[width=0.80\linewidth]{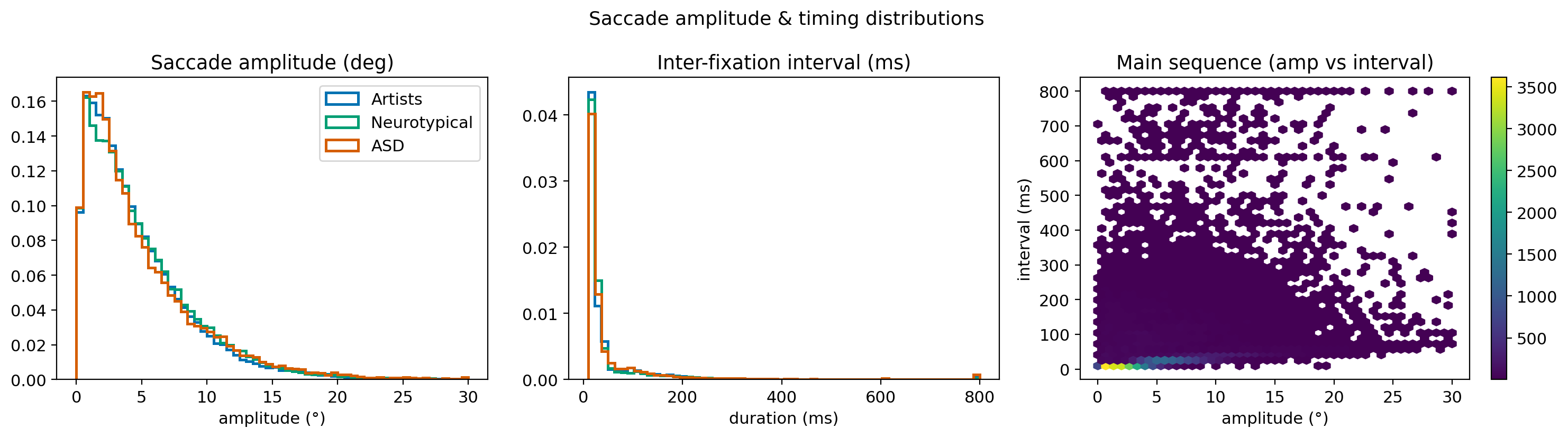}
  \caption{Saccade amplitude (deg) and inter-fixation interval (ms) distributions
  by group, and the amplitude--interval main sequence (2-D density).}
  \label{fig:s-mainseq}
\end{figure}

\begin{figure}[htbp]
  \centering
  \includegraphics[width=0.58\linewidth]{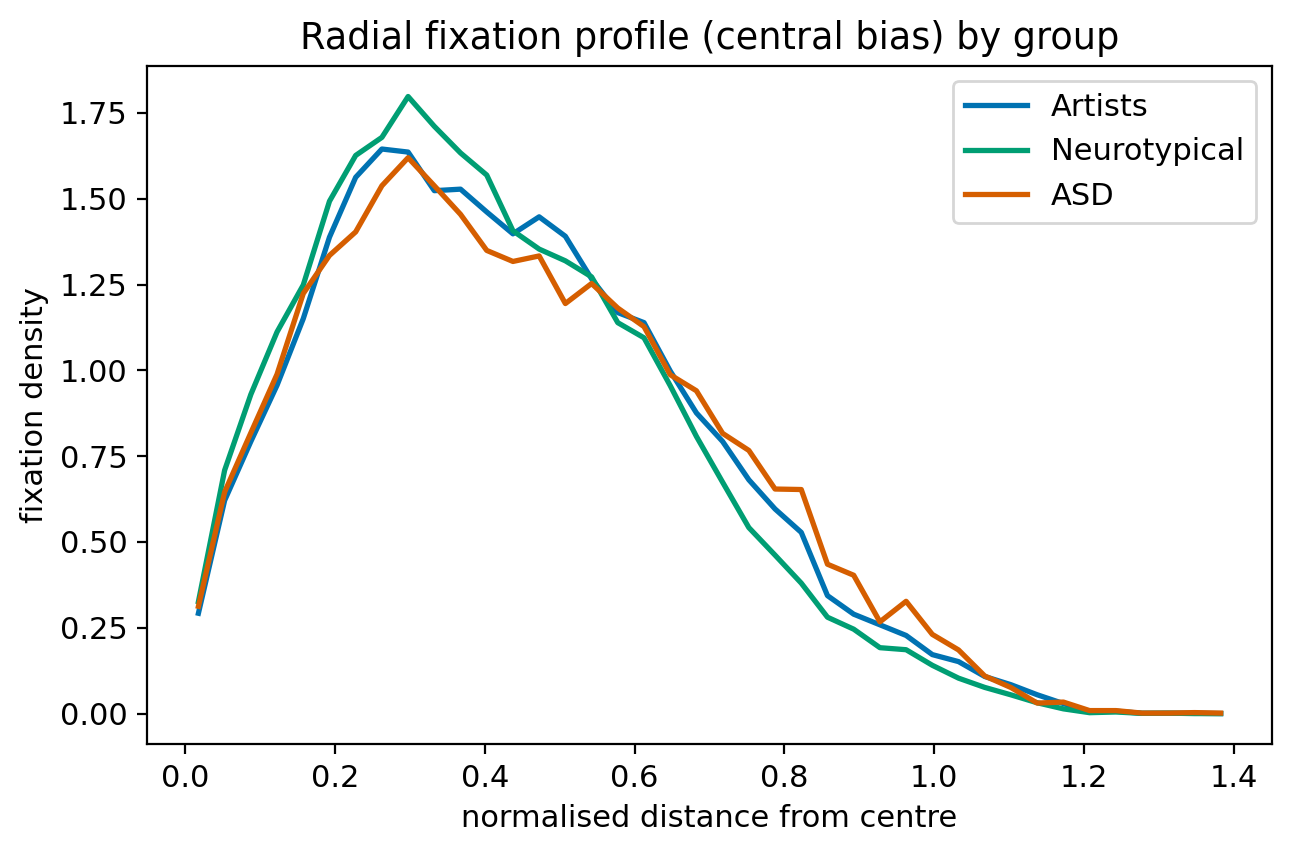}
  \caption{Radial fixation profile: fixation density as a function of normalised
  distance from the screen centre, by group. ASD is shifted outward.}
  \label{fig:s-radial}
\end{figure}

\begin{figure}[htbp]
  \centering
  \includegraphics[width=0.80\linewidth]{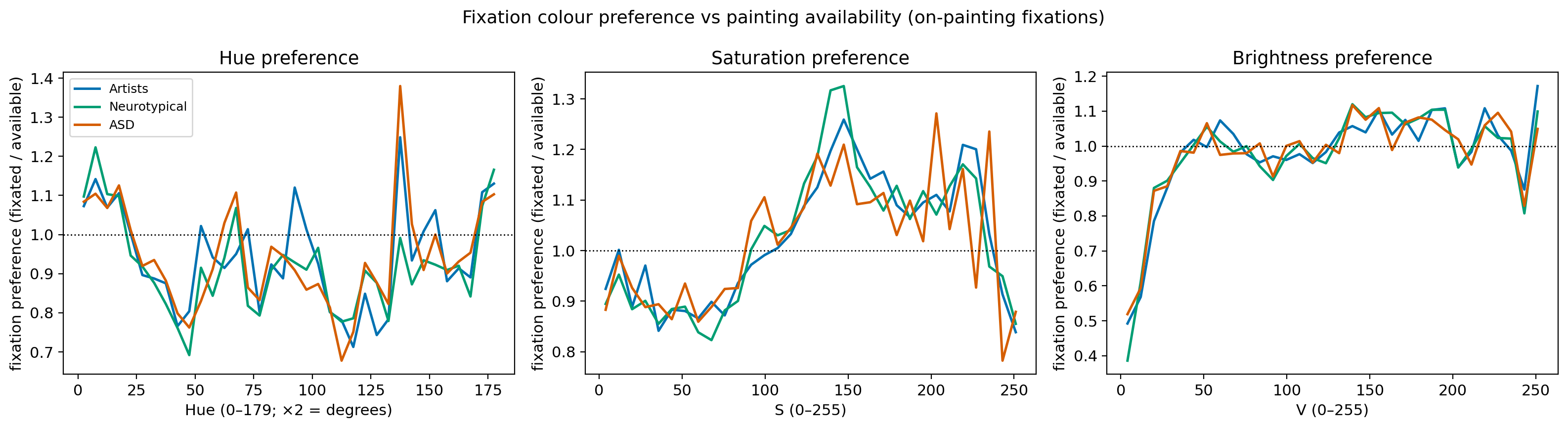}
  \caption{Fixation colour preference (fixated density / available density) by
  hue, saturation and brightness, per group. Values above~1 indicate a band
  fixated more than its areal availability predicts.}
  \label{fig:s-huepref}
\end{figure}

\begin{figure}[htbp]
  \centering
  \includegraphics[width=0.58\linewidth]{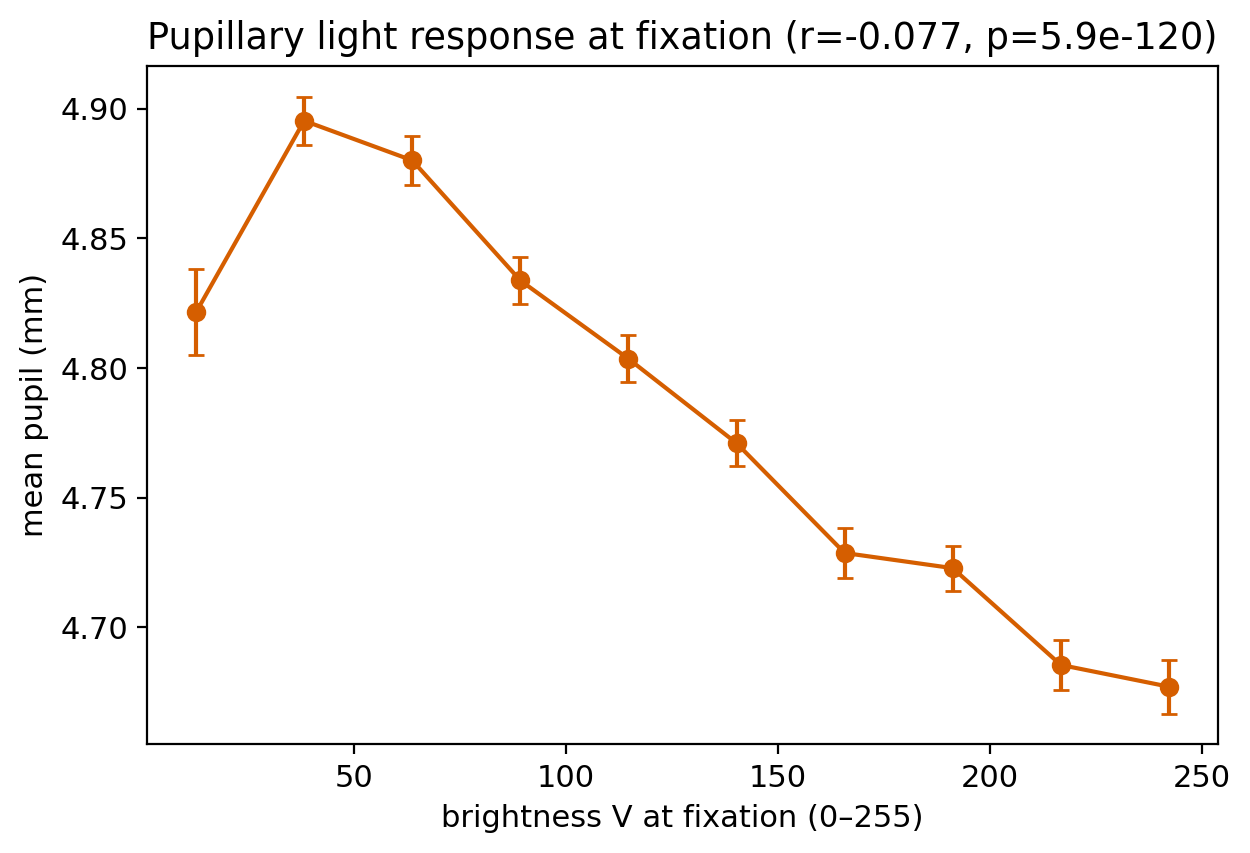}
  \caption{Pooled pupillary light response: mean pupil diameter as a function of
  the brightness at the fixated location.}
  \label{fig:s-pupilbright}
\end{figure}

\section{Software and reproducibility}
Analyses were run in Python 3.12 with: \texttt{eyefeatures} 2.0.2 (IDT
extraction), \texttt{scanpath-nlp-metrics} 0.0.1 (MultiMatch, ScanMatch, DTW),
\texttt{fdiss} 0.1.0 (FDISS), \texttt{numpy} 1.26, \texttt{pandas} 2.2,
\texttt{scipy} 1.14, \texttt{statsmodels} 0.14, \texttt{opencv} 4.11, and
\texttt{matplotlib} 3.8. The pipeline (fixation extraction, spatial metrics,
scanpath comparison, feature extraction, statistics, figures) and all
per-stimulus outputs are released at
\url{https://github.com/kmamine/TSA-Art-Typ-results}.

\end{document}